\documentclass[twoside]{article}

\usepackage[accepted]{aistats2025}

\usepackage[round]{natbib}

\usepackage{bm}
\usepackage{amssymb}
\usepackage{amsmath, amsthm}
\usepackage{algorithm}
\usepackage{algpseudocode}
\usepackage{graphicx} 
\usepackage{xcolor}
\usepackage{subcaption}  
\usepackage[normalem]{ulem}
\usepackage{dsfont}
\usepackage{multirow} 
\usepackage{url}
\usepackage[hyphenbreaks]{breakurl}

\newtheorem{definition}{Definition}
\newtheorem{theorem}{Theorem}
\newtheorem{lem}{Lemma}

\newenvironment{customproof}[1][Proof]{\noindent\textbf{#1.} }{\qed \vspace{10pt}}
\newcommand*\diff{\mathop{}\!\mathrm{d}}

\newcommand{\indicate}[1]{\mathds{1}{[#1]}}

\begin{document}

%

%

\runningauthor{Qingshi Sun, Nathan Justin, Andr\'es G\'omez, Phebe Vayanos}

\twocolumn[

\aistatstitle{Mixed-feature Logistic Regression Robust to Distribution Shifts}

\aistatsauthor{ Qingshi Sun$^{1,3}$  \And Nathan Justin$^{2,3}$  \And  Andr\'es G\'omez$^{1,3}$ \And Phebe Vayanos$^{1,2,3}$ }

\aistatsaddress{\textsuperscript{\rm 1}Department of Industrial \& Systems Engineering, USC \\ 
\textsuperscript{\rm 2}Department of Computer Science, USC \\ 
\textsuperscript{\rm 3}Center for Artificial Intelligence in Society, USC \\
 \{qingshis,njustin,gomezand,phebe.vayanos\}@usc.edu
} ]

\begin{abstract}
Logistic regression models are widely used in the social and behavioral sciences and in high-stakes domains, due to their simplicity and interpretability properties. At the same time, such domains are permeated by distribution shifts, where the distribution generating the data changes between training and deployment. In this paper, we study a distributionally robust logistic regression problem that seeks the model that will perform best against adversarial realizations of the data distribution drawn from a suitably constructed \emph{Wasserstein ambiguity set}.  Our model and solution approach differ from prior work in that we can capture settings where the likelihood of distribution shifts can vary across features, significantly broadening the applicability of our model relative to the state-of-the-art.  We propose a graph-based solution approach that can be integrated into off-the-shelf optimization solvers. We evaluate the performance of our model and algorithms on numerous publicly available datasets. Our solution achieves a 408x speed-up relative to the state-of-the-art. Additionally, compared to the state-of-the-art, our model reduces average calibration error by up to 36.19\% and worst-case calibration error by up to 41.70\%, while increasing the average area under the ROC curve (AUC) by up to 18.02\% and worst-case AUC by up to 48.37\%.
\end{abstract}

\section{INTRODUCTION}

Machine learning plays a critical role in decision-making in high-stake domains, such as healthcare, social science, finance, and criminal justice \citep{simplemodel}. Within such domains, model transparency and interpretability are critical to ensure trustworthiness~\citep{inte10}. One of the most commonly used classification methods in such settings is classical logistic regression, which models the probability of a binary outcome by mapping input features to a probability value using a logistic function~\citep{logisticregression}. Logistic regression is highly interpretable as it models the log-odds of an event as a linear combination of the covariates, making it easy to understand how changes in input variables affect the probability outcome.

However, in high-stakes domains where this model is often used, distribution shifts are ubiquitous. These refer to differences between the training data distribution and the data encountered during testing or deployment. 
In this work, we specifically address conditional shift, where the conditional distribution of the input features given the label varies, while the marginal distribution of the label remains unchanged \citep{conditionalshift}. This phenomenon can also be viewed as a form of concept drift \citep{ConceptDriftReview, ConceptDriftSurvey}. 
 For example, in the high-stakes domain of public housing allocation, personal information used to predict homelessness risk is often collected through self-reported surveys \citep{CESTTRR2023}. Over time, refinements in data collection practices—such as changes in question phrasing or survey locations—can lead to shifts in the way individual features are reported, even when an individual’s homelessness status (label) remains unchanged. At the same time, the distribution of the label remains consistent because the definition of homelessness is independent of how the personal information is collected. Conditional shifts can occur due to changes in data collection protocols, technological advancements, or environmental variations, often leading to a significant decline in model performance.
Similarly to most machine learning (ML) models, logistic regression is susceptible to distribution shifts. It is therefore critical to develop logistic regression models that remain \emph{robust}, even in the presence of changing data distributions.

In recent years, distributionally robust optimization (DRO) methods have been introduced to enhance the robustness of ML models against distribution shifts~\citep{aos, labelshift, groupdistributionally}. In ML, DRO seeks to find the model that will perform best in the worst-case realization of the distribution of the data within a suitably constructed ambiguity set that captures e.g., prior knowledge about the distribution of the data and about the likelihood of different shifts.

In the literature, many metrics are proposed to construct ambiguity sets, such as moment based uncertainty~\citep{DY10:distr_rob_opt}, Kullback-Leibler divergence~\citep{L19:kl_divergence}, and Wasserstein distance~\citep{datadrivenDRO}. In particular, \citet{datadrivenDRO} have shown that DRO problems based on the Wasserstein distance can be reformulated as finite convex programs under mild assumptions and that they have attractive convergence properties and provable out-of-sample performance guarantees.

In our work, we study a distributionally robust variant of logistic regression where the distance between the training data distribution and the testing/deployment data distribution is measured by the Wasserstein metric. Our work most closely relates to two papers in the literature. \citet{NIPS2015} are the first to propose a DRO approach to logistic regression, showing that a distributionally robust logistic regression problem admits an equivalent reformulation as a polynomial-size convex optimization problem if all features are numerical. Then, \citet{closest} extend this approach to solve the problem when both numerical and categorical features are present, resulting in an exponential-size convex optimization problem. They also propose a cutting-plane method to solve the problem as a sequence of polynomial-time solvable programs.  Importantly, both of these works assume that all features are equally prone to shift, i.e., that the likelihood of a shift is equally likely across all features.

However, in real-world situations, distribution shifts often affect different features differently and we have access to partial information about the likelihood of different shifts; for example, we may know that certain features are more prone to variability than others. 
Prior work has explored the design of Wasserstein ambiguity sets to account for feature heterogeneity. \citet{CostSelection} propose a Mahalanobis-based distance metric to incorporate feature heterogeneity; however, their approach does not explicitly address parameter selection under distribution shifts. Building on this work, \citet{RethinkingDistributionShifts} introduce a heuristic approach that prioritizes features with substantial shifts. However, their method is specifically designed for covariate shifts \citep{distributionshift}, assumes access to target domain data, and only considers a binary (0-1) weighting scheme when modeling feature heterogeneity. In many real-world scenarios, access to target domain data cannot be guaranteed. Furthermore, a rigid binary weighting scheme lacks flexibility in modeling varying degrees of distribution shifts across features.

In this paper, we study the problem of learning a logistic regression model that is robust to conditional shifts where the likelihood of a shift may differ across features  and applicable to datasets involving mixed features. The goal is to ensure the best possible performance under conditional shifts in the training data, leading to reliable outcomes during deployment. The main contributions of our work are:
\begin{itemize}
    \item On the model side, we propose a distributionally robust logistic regression model with a Wasserstein ambiguity set that accounts for distribution shifts, where such shifts can affect different features differently. Additionally, we introduce a fine-grained method for calibrating the ambiguity set based on basic domain knowledge.
    
    \item On the algorithmic side, we adapt the cutting-plane method in \citet{closest} to solve our proposed model. Additionally, we develop a graph-based reformulation that significantly reduces runtime relative to the cutting-plane method, making it practically scalable for training models on large datasets. 
    
    \item On the computational side, we demonstrate that our proposed distributionally robust logistic regression improves performance across several metrics, including calibration error and AUC, compared to both standard regularized logistic regression and existing models under distribution shifts. In particular, compared to the state-of-the-art model, our model reduces average calibration error by up to 36.19\% and worst-case calibration error by up to 41.70\%, while increasing the average AUC by up to 18.02\% and worst-case AUC by up to 48.37\%.
    Moreover, the graph-based formulation can be solved up to 408.12 times faster on average compared to the state-of-the-art algorithm.
\end{itemize}

\textbf{Notation.} We define $[N] = \{ 1, \ldots, N \}$ for $N \in \mathbb{N}$. The indicator function, denoted as $\indicate{\mathcal{E}}$, equals $1$ if the condition $\mathcal{E}$ is satisfied and $0$ otherwise. The set of all probability distributions supported on a set $\Xi$ is denoted by $\mathcal{P}_0 (\Xi)$.

\section{DISTRIBUTIONALLY ROBUST OPTIMIZATION FORMULATION}
\label{sec:formulation}

\subsection{Wasserstein Logistic Regression and Ambiguity Set} 
\label{sec:notation and problem def}

We consider a mixed-feature classification dataset with $N$ data points, $\{\bm{\xi}^i := (\bm{x}^i, \bm{z}^i,  y^i)\}_{i \in [N]}$, where $\bm{\xi}^i$ collects the features and labels related to datapoint~$i$. Specifically, each datapoint $i$ is characterized by~$n$ numerical features $\bm{x}^i = (x^i_{1}, \ldots, x^i_{n}) \in \mathbb{R}^n$, $m$ categorical features $\bm{z}^i = (\bm{z}^i_1, \ldots, \bm{z}^i_m) \in \mathcal{C}_1 \times \ldots \times \mathcal{C}_m$, and a binary label $y^i \in \{ -1, + 1 \}$. Here, for $j \in [m]$, if the number of  possible values of $\bm{z}_j$ is $a$, $\mathcal{C}_j := \{ \bm{z} \in \{0,1\}^{a - 1} \, : \, \sum_{k \in [a - 1]} z_k \leq 1 \}$,  representing the set of one-hot encoded categorical features associated with feature $j$. 
We denote by $\mathcal{C} := \mathcal{C}_1 \times \ldots \times \mathcal{C}_m$ and $\Xi := \mathbb{R}^n \times \mathcal{C} \times \{ -1, +1 \}$ the support of the categorical features as well as the dataset, respectively, and we let~$c$ be the number of encoded categorical features.

We select Wasserstein ball $ \mathfrak{B}_{\epsilon}(\widehat{\mathbb{P}}_{N})$ as the ambiguity set, which contains all possible distributions  $\mathbb{Q}$ centered at the empirical distribution $\widehat{\mathbb{P}}_N$. Since the true distribution of the training data is unknown in practice, we use the empirical distribution $\widehat{\mathbb{P}}_N := \frac{1}{N} \sum_{i = 1}^N \delta_{\bm{\xi}^i}$ that places equal probability mass on all data points $\{ \bm{\xi}^i \}_{i \in [N]}$, where $ \delta_{\bm{\xi}^i}$ denotes the Dirac point measure at $\bm{\xi}^i$. We aim to solve the following logistic regression problem robust to distribution shifts:
\begin{equation}
\begin{array}{cl}
\underset{\boldsymbol{\beta}}{\operatorname{minimize}} & \displaystyle \sup _{\mathbb{Q} \in \mathfrak{B}_{\epsilon}\left(\widehat{\mathbb{P}}_{N}\right)} \mathbb{E}_{\mathbb{Q}}\left[l_{\bm{\beta}}(\bm{x}, \bm{z}, y)\right] \\
\text {subject to} & \boldsymbol{\beta}=\left(\beta_{0}, \bm{\beta}_{\rm x}, \bm{\beta}_{\rm z} \right) \in \mathbb{R}^{1+n+c},
\end{array}
\label{eq:original formulation}
\end{equation}
where $l_{\bm{\beta}}$ is the log-loss function defined through
\[
l_{\bm{\beta}}(\boldsymbol{x}, \boldsymbol{z}, y)
:= \log \left( 1 + \exp \left( -y \cdot \left( \beta_{0} 
+ \boldsymbol{\beta}_{\rm x}^{\top} \boldsymbol{x}
+ \bm{\beta}_{\rm z}^{\top} \bm{z} \right) \right) \right).
\]
Problem \eqref{eq:original formulation} hedges against a broad range of potential distributions and identifies the optimal coefficients by minimizing the expected log loss under the worst-case distribution, providing a safeguard against data deviations during deployment.

Formally, we define 
\begin{equation}\label{eq:ambiguity set}
   \mathfrak{B}_\epsilon(\hat{\mathbb{P}}_N):=\{\mathbb{Q} \in \mathcal{P}_0(\Xi) : W(\mathbb{Q},\hat{\mathbb{P}}_N) \leq \epsilon\} 
\end{equation}
 as the ball of the radius $\epsilon \in \mathbb{R}^+$ centered at $\hat{\mathbb{P}}_N$ with respect to the Wasserstein distance defined below.

\begin{definition}[Wasserstein Distance]
\textnormal{The Wasserstein distance between two distributions $\mathbb{P}$ and $\mathbb{Q}$ supported on $\Xi$ is defined as}
\begin{equation}
    \begin{array}{l}
    W(\mathbb{Q},\mathbb{P}) \; := \; \displaystyle \inf_{\Pi\in \mathcal{P}_0(\Xi^2)} \Bigl\{\int_{\Xi^2} d(\boldsymbol{\xi},\boldsymbol{\xi}^{\prime}) \Pi({\textnormal{d}}\boldsymbol{\xi},\textnormal{d}\boldsymbol{\xi^{\prime}}) : \notag \\
    \quad \Pi(\textnormal{d}\boldsymbol{\xi},\Xi) = \mathbb{Q}(\textnormal{d}\boldsymbol{\xi}), \Pi(\Xi, \textnormal{d}\boldsymbol{\xi^{\prime}}) = \mathbb{P}(\textnormal{d}\boldsymbol{\xi^{\prime}})\Bigr\},
    \end{array}
\end{equation}
 \textnormal{where $\bm{\xi} = (\bm{x},\bm{z},y) \in \Xi$ and $\bm{\xi}' = (\bm{x}',\bm{z}',y') \in \Xi$, while $d(\bm{\xi},\bm{\xi}')$ is a weighted distance metric on $\Xi$.}
\end{definition}
The Wasserstein distance represents the minimum cost of moving the distribution $\mathbb{P}$ to the distribution $\mathbb{Q}$, where the cost of moving a unit mass from $\bm{\xi}$ to $\bm{\xi}'$ amounts to $d(\bm{\xi},\bm{\xi}')$. 
	
Next, we define the weighted distance metric used in the Wasserstein distance.  
\begin{definition}[Weighted Distance Metric] \label{def:metric}
\textnormal{We measure the distance between two data points $\bm{\xi} = (\bm{x},\bm{z},y) \in \Xi$ and $\bm{\xi}'=(\bm{x}',\bm{z}',y') \in \Xi$ as}
\begin{equation*}
    \begin{array}{l}
    d (\bm{\xi},\bm{\xi}') \; := \; \displaystyle \sum_{j \in [n]} \gamma_{j} |x_j - x'_j| +  \displaystyle \sum_{\ell \in [m]} {\delta_{\ell}} \indicate{\bm{z}_{\ell} \neq \bm{z}'_{\ell}}  \\
      \qquad \qquad \qquad \qquad
        + \kappa \cdot \indicate{y \neq y'}
    \end{array}
\end{equation*}
\textnormal{for some $\gamma_{j}>0$, $\delta_{\ell}>0$, and $\kappa >0$.}
\end{definition}
In this definition, the weighted distance between numerical features $\bm{x}$ and $\bm{x}'$ is defined by the norm-based difference. Here, $\gamma_{j}$ is the weight of the perturbation of numerical feature $j$, representing the cost per unit difference between $x_j$ and $x'_j$. The weighted distance between two categorical feature vectors $\bm{z}$ and $\bm{z}'$ is defined by the discrepancies between their corresponding components.  $\delta_{\ell}$ represents the weight associated with the perturbation of categorical feature~${\ell}$, representing the cost when $\bm{z}_{\ell}$ and $\bm{z}'_{\ell}$ differ.
 Similarly, the discrepancy between the labels $y$ and $y'$ is accounted for by a constant $\kappa$.
By selecting proper weights, we can account for the relative sensitivity of the model to different features, ensuring that the optimization is more robust to shifts in critical dimensions while reducing unnecessary conservatism in others. In this work, we use weight parameters \( \gamma_{j} \) and \( \delta_{\ell} \) in the ambiguity set to account for the varying likelihood of distribution shifts across different features. 
We discuss a method for calibrating all parameters in the ambiguity set \eqref{eq:ambiguity set} in Section \ref{sec:calibration}.

\subsection{Reformulation as a Convex Problem with Exponential Number of Constraints}

In this work, since the marginal distribution of label in the testing data remains unchanged, we set~\(\kappa\) in Definition~\ref{def:metric} to infinity, implying that any shifts in the labels will result in distributions falling outside the ambiguity set.
\begin{theorem}[Convex Formulation] \label{convex formulation} Without shifts in labels, the distributionally robust logistic regression problem \eqref{eq:original formulation} can be reformulated as 
\begin{equation} \label{eq:convex reformulation}
\begin{array}{rl}
\underset{\lambda, \boldsymbol{r}, \bm{\beta}}{\text{\emph{min}}} & \lambda \epsilon + \frac{1}{N} \sum_{i \in [N]} r_i \\[1.2em]
\text{\emph{s.t.}} & l_{\bm{\beta}} (\bm{x}^i, \bm{z}, y^i) - \lambda \sum_{\ell \in [m]} \delta_{\ell} \indicate{\bm{z}_{\ell} \neq \bm{z}^i_{\ell}} \leq r_i, \\[1em]
& \qquad \qquad \quad \; \forall i \in [N], \; \bm{z} \in \mathcal{C} \\[1em]
&  |\gamma_{j}^{-1} \beta_{\mathrm{x}j}| \leq \lambda, \; \forall j \in [n] \\[1em]
& \lambda \geq 0, \; \boldsymbol{r} \in \mathbb{R}^N, \; {\bm \beta} = (\beta_0, \bm{\beta}_{\mathrm{x}}, \bm{\beta}_{\mathrm{z}}) \in \mathbb{R}^{1+n+c}
\end{array}
\end{equation}
where $\lambda$ and $\bm{r}$ are dual variables arising by dualizing the inner maximization problem in~\eqref{eq:original formulation}.  
The constraints with log-loss functions can be converted into exponential cone format, resulting in a convex problem that can be solved with off-the-shelf solvers.
\end{theorem}
The proof of Theorem \ref{convex formulation} and the equivalent formulation of problem \eqref{eq:convex reformulation} to an exponential cone problem are given in Appendix~\ref{sec:proof of convex formulation}. While problem~\eqref{eq:convex reformulation} can in principle be solved with off-the-shelf solvers, it contains exponentially many constraints, making it impractical to solve monolithically when the number of possible realizations of each categorical feature is large. In Section \ref{sec:solution}, we propose two solution methods that will enable us to solve practically sized problems.

\section{PARAMETER CALIBRATION} \label{sec:calibration}
We develop a fine-grained method for calibrating the ambiguity set radius $\epsilon$ and the weight parameters $\gamma_{j}$ and $\delta_{\ell}$ where $j \in [n]$ and $\ell \in [m]$ based on the estimated likelihood of distribution shifts.
We build on the calibration method proposed by \citet{decisiontree} in the context of robust optimization with discrete features. We adapt this technique to the calibration of parameters in our \emph{distributionally} robust optimization problem, and extend its application from discrete features to numerical features.

Without access to target domain data, we rely on basic domain knowledge to characterize uncertainty in numerical features. Specifically, for each numerical feature \( j \in [n] \), we denote the corresponding feature after distribution shifts as \( \tilde{x}_j \) and define its probability density function \( f \) given \( x_j \). To quantify uncertainty, we assume a probability of certainty \( \rho_{\mathrm{x}j} \in (0,1] \) and an interval \( [l_j, u_j] \), where 

\begin{equation} \label{eq:numerical prob of certainty relationship}
\rho_{\mathrm{x}j} = \mathbb{P}(l_j \leq \tilde{x}_j - x_j \leq u_j) = \int_{l_j + x_j}^{u_j + x_j} f(\Tilde{x}_j) \text{d}\Tilde{x}_j.
\end{equation}

This represents the probability that the shift in \( x_j \) during deployment falls within the interval \([l_j, u_j]\). These intervals and probabilities can be estimated from historical data or expert evaluation.
A feature with a low probability of certainty and a large interval is highly susceptible to shifts, whereas a feature with a high probability of certainty and a small interval is more stable and likely to retain a similar distribution in deployment as observed during training.

In the absence of further knowledge about the distribution shift, we follow the principle of maximum entropy~\citep{entropy}, which chooses the distribution of the perturbations with greatest entropy and thus highest uncertainty subject to our assumption of the probability of certainty $\rho_{\mathrm{x}j}$ and $\rho_{\mathrm{z}\ell}$.
To this end, we select the Laplace distribution to reflect shifts on the numerical features. In particular, we assume that the shift of numerical feature $j$ follow a Laplace distribution with scale parameter~$b_j$ and location parameter 0. We also assume that the shift does not have a specific trend in the positive or negative direction and thus set $u_j = - l_j$. Specifically, given \( x_j \), the probability density function $f$ of $\tilde{x}_j$ is defined as
\begin{equation} \label{eq:numerical pdf}
   f(\tilde{x}_j
) = \frac{1}{2b_j}\exp{\left(-\frac{\lvert \tilde{x}_j - x_j
 \rvert}{b_j}\right)}.
\end{equation}

Combining \eqref{eq:numerical prob of certainty relationship} and \eqref{eq:numerical pdf}, we obtain $b_j = \frac{-u_j}{\log(1-\rho_{\mathrm{x}j})}$.

Similarly,  for each categorical feature \( \ell \in [m] \), we denote the probability of~$\bm{z}_{\ell}$ remaining unchanged in deployment as \( \rho_{\mathrm{z}\ell} \in \left[\frac{1}{|\mathcal{C}_{\ell}|}, 1\right]\). Following the principle of maximum entropy, we assume that the probability of perturbing~$\bm{z}_{\ell}$ to~$\tilde{\bm{z}}_{\ell} \in \mathcal{C}_{\ell}$ is
\begin{equation} \label{eq:categorical pmf}
    \mathbb{P}(\tilde{\bm{z}}_{\ell}) = \indicate{\tilde{\bm{z}}_{\ell} = \bm{z}_{\ell}} \rho_{\mathrm{z}\ell} + \indicate{\tilde{\bm{z}}_{\ell} \neq \bm{z}_{\ell}} \frac{1 - \rho_{\mathrm{z}\ell}}{ |\mathcal{C}_{\ell}| - 1}.
\end{equation}
The above probability mass function allows for any perturbation of $\bm{z}_{\ell}$ to occur with the same probability~$\frac{1 - \rho_{\mathrm{z}\ell}}{ |\mathcal{C}_{\ell}| - 1}$. 

Lastly, we adopt the approach of constructing ambiguity sets using hypothesis testing. Specifically, we perform a likelihood ratio test on the magnitude of the perturbation, with a threshold determined by \( \theta \), where \( \theta \in (0,1] \) controls the level of robustness. A value of~$\theta$ close to 0 signifies a fully robust model and~$\theta=1$ signifies no robustness in the model.
We refer to Appendix~\ref{sec:details of calibration} on how the probability density function~\eqref{eq:numerical pdf}, probability mass function~\eqref{eq:categorical pmf}, and the value of $\theta$ can be used to calibrate the values of $\gamma_j$, $\delta_{\ell}$, and $\epsilon$, respectively, through hypothesis testing. From our derivations in Appendix~\ref{sec:details of calibration}, we set~$\gamma_{j} = \frac{-\log(1 - \rho_{\mathrm{x}j})}{u_j}$,~$\delta_{\ell} = \log \left( \frac{\rho_{\mathrm{z}\ell} (|\mathcal{C}_{\ell}| - 1)}{1 - \rho_{\mathrm{z}\ell}} \right)$, and~$\epsilon = -\log \theta$ as the tuned parameters of our ambiguity set.

Notably, by using application-specific information, alternative distributions of perturbations can be tailored to more accurately capture the distribution shifts. Our calibration method adjusts the parameters $\gamma_j$, $\delta_\ell$, and $\epsilon$ accordingly. Although this method involves domain knowledge, the numerical results in Appendix \ref{sec:Unexpected Perturbations} show that even with imprecise estimates and model misspecification, our calibrated model maintains strong performance under distribution shifts.

\section{SOLUTION METHODS}\label{sec:solution}

\subsection{Cutting-Plane Method} \label{sec:cutting method}
We adapt the cutting-plane method from \citet{closest} to solve the distributionally robust logistic regression problem \eqref{eq:convex reformulation} with ambiguity set~\eqref{eq:ambiguity set}. At each iteration, the algorithm solves a relaxed version of problem \eqref{eq:convex reformulation}, identifies the constraints in the original problem that are violated the most by the current solution to the relaxed problem, and incorporates these constraints into the relaxed problem to progressively improve the solution. This process continues until no violated constraints remain, ensuring convergence to the optimal solution. Our proposed cutting-plane approach differs from that in \citet{closest} in the method we use to identify violated constraints. Indeed, the approach from \citet{closest} does not apply when the weight parameters $\delta_{\ell}$ are not all 1. Their constraint identification algorithm relies on a direct mapping between the distance $d (\bm{\xi},\bm{\xi}')$ and the number of feature disagreements. When weights vary, this relationship breaks, making it difficult to determine which features differ or how many are different.
In what follows we focus on describing our method for identifying violated constraints. For details on how this procedure can be integrated in a cutting-plane algorithm for solving problem \eqref{eq:convex reformulation}, we refer to Appendix~\ref{sec:details of cutting-plane scheme}.

We now describe our procedure to identify, for any fixed solution $(\lambda,\bm{r}, \bm{\beta})$ to the relaxed problem, the most violated constraints in problem \eqref{eq:convex reformulation}. These are  indexed by $(i,\bm{z}) \in [N] \times \mathcal{C}$.
We propose to solve, for each $i \in [N]$, the following optimization problem:
\begin{equation}\label{eq:identification}
\begin{aligned}
    \underset{\bm{z} \in \mathcal{C}}{\text{maximize}} \ & \log \left( 1 + \exp \left( -y^i \left( \bm{\beta}_{\mathrm{x}}^\top \bm{x}^i + \bm{\beta}_{\mathrm{z}}^\top \bm{z} + \beta_0 \right) \right) \right) \\
    & \quad - \lambda \sum_{\ell \in [m]} \delta_{\ell} \indicate{\bm{z}_{\ell} \neq \bm{z}^i_{\ell}}  -r_i.
\end{aligned}
\end{equation}
In this problem, any solution $\bm{z} \in \mathcal{C}$ that results in an objective value greater than~0 corresponds to a violated constraint. 
Problem \eqref{eq:identification} is challenging due to the exponential number of possible realizations of $\bm{z}$. In problem \eqref{eq:identification}, the decision variables $\bm{z}$ only appear in the linear term \( -y^i \sum_{\ell \in [m]}\bm{\beta}_{\mathrm{z}\ell}^\top \bm{z}_{\ell}\) and the weighted distance $\sum_{\ell \in [m]} \delta_{\ell} \indicate{\bm{z}_{\ell} \neq \bm{z}^i_{\ell}}$. At the same time, since both terms are linearly separable, this structure allows for the decomposition of the problem into subproblems, where each subproblem considers a subset of categorical features and fixes a weighted distance corresponding to those features. Therefore, we propose to solve this problem using dynamic programming, which allows us to decompose the problem into simpler subproblems, conditioned on weighted distances, to speed-up computation.

For any fixed datapoint~$i$, we define the set of non-root and non-terminal states of our dynamic program as  
\begin{equation*}
\begin{aligned}
    \mathcal{S}^i_1 := & \left\{ (k,d) \mid d = \sum_{\ell=1}^{k} \delta_{\ell} \indicate{\bm{z}_{\ell} \neq \bm{z}^i_{\ell}}, \; k \in [m], \; \bm{z} \in \mathcal{C} \right\}.
\end{aligned}
\end{equation*}
In this definition, $k \in [m]$ specifies that categorical features indexed from 1 to $k$ are considered, and $d$ represents the sum of the weighted distances between a value $\bm{z}_{\ell} \in \mathcal{C}_{\ell}$ and $\bm{z}_{\ell}^i$ in the dataset across all categorical features $\ell \in [k]$.  
We also define a set $\mathcal{S}_0$ containing the root state and terminal state:
$ \mathcal{S}_0 = \left\{ (0,0), (m+1,0) \right\}$. 
The state space of each dynamic programming problem indexed by data point $i$ is $\mathcal{S}^i = \mathcal{S}^i_1 \cup \mathcal{S}_0$.
We denote the optimal objective value of the dynamic programming subproblems by a function \( g^i: \mathcal{S}^i \rightarrow \mathbb{R} \). That is, for each state \( (k,d) \in \mathcal{S}^i_1 \), we define
\begin{equation}\label{eq:subproblem k=1m}
    \begin{array}{ccl}
        g^i(k,d) \; := & \underset{\{\bm{z}_{\ell}\}_{\ell \in [k]}}{\text{max}} &  \displaystyle  -y^i \sum_{\ell \in [k]} \bm{\beta}_{\mathrm{z}\ell}^\top \bm{z}_{\ell}   \\
        & \text{s.t.}  & {\bm{z}_{\ell} \in \mathcal{C}_{\ell}, \; \ell \in [k]} \\
        && \displaystyle \sum_{\ell \in [k]} \delta_{\ell} \indicate{\bm{z}_{\ell} \neq \bm{z}^i_{\ell}} = d\\
    \end{array}
    \end{equation}

Each subproblem \eqref{eq:subproblem k=1m} only considers categorical features indexed from $1$ to $k$ for $k \in [m]$ and is conditioned on the weighted distance $d$ corresponding to those $k$ categorical features. 
Additionally, we set $g^i(0,0): =0$ and denote by $g^i(m+1,0)$ the optimal objective value of problem~\eqref{eq:identification}.

Subproblems~\eqref{eq:subproblem k=1m} can be solved recursively using the following Bellman equations.

If $k \in [m]$, 
\begin{equation}\label{eq:bellman g k=1m}
\begin{array}{cl}
    & g^i(k,d) \; = \;  \underset{\bm{z}_{k} \in \mathcal{F}_{kd}}{\text{max}}  -y^i \bm{\beta}_{\mathrm{z}k}^\top \bm{z}_{k} + g^i\left(k-1, \right. \\
   & \qquad \qquad \qquad \qquad \qquad \left. d - \delta_{k} \indicate{\bm{z}_{k} \neq \bm{z}^i_{k}}\right), 
\end{array}
\end{equation}
where
\[
\mathcal{F}_{kd} := \left\{ \bm{z}_k \in \mathcal{C}_k \mid 
(k-1, d - \delta_{k} \indicate{\bm{z}_k \neq \bm{z}^i_k})
\in \mathcal{S}^i \right\}.
\]
If $k = m+1$, 
\begin{equation}\label{eq:bellman g k=m+1}
\begin{aligned}
 g^i(m+1, 0)  \; = \; & \underset{d}{\text{max}} \; \log\left(1 + \exp\left(-y^i \bm{\beta}_{\mathrm{x}}^\top \bm{x}^i \right.\right. \\
    & \qquad  \; \left.\left. + g^i(m, d)\right)\right) - \lambda d - r_i \\
    & \text{s.t.} \quad (m, d) \in \mathcal{S}^i_1.
\end{aligned}
\end{equation}

Starting from the simplest case with a single categorical feature, Bellman equation \eqref{eq:bellman g k=1m} incrementally solves subproblem \eqref{eq:subproblem k=1m} by adding one categorical feature at a time. The linear separability of the objective function 
$- y^i \sum_{\ell \in [k]} \boldsymbol{\beta}_{\mathrm{z}\ell}^\top \boldsymbol{z}_\ell$ 
enables this incremental approach. Bellman equation \eqref{eq:bellman g k=1m} follows this strategy by leveraging the optimal solution of subproblem \eqref{eq:subproblem k=1m} with categorical features indexed from \( 1 \) to \( k-1 \) and evaluating all feasible values of the \( k \)th categorical feature. Finally, solving subproblem \eqref{eq:subproblem k=1m} for \( k = m \) and a given distance \( d \) for $(m,d) \in \mathcal{S}^i$ yields \( g^i(m, d) \).
Bellman equation \eqref{eq:bellman g k=m+1} compares $g^i(m,d)$ with the corresponding solution for all $(m,d) \in \mathcal{S}^i$ to obtain the optimal solution for problem \eqref{eq:identification}.
Algorithm \ref{alg:dp} provides the details of this dynamic programming procedure. The proof of correctness of Algorithm~\ref{alg:dp} can be found in Appendix~\ref{sec: proof of dp algorithm}.

\begin{algorithm}[tb]
\caption{Identification of Most Violated Constraints in Problem~\eqref{eq:convex reformulation} given the data point indexed by $i$.} 
    \label{alg:dp}
    \begin{algorithmic}[1]  
        \Statex \textbf{Input}: A solution $(\lambda, \bm{r}, \bm{\beta})$ to the relaxed problem of \eqref{eq:convex reformulation} and a datapoint $i \in \Xi$
        \Statex \textbf{Output}: optimal value to problem \eqref{eq:identification} and a corresponding optimal solution.

        \State \textbf{Initialize} $\bm{z}_{[0]}(0) \leftarrow ()$       
        \State \textbf{Initialize} $g^i(0,0) \leftarrow 0$
        \For{$(k,d) \in \mathcal{S}^i_1$}
            \State \textbf{Solve} $g^i(k,d)$ by Bellman equation \eqref{eq:bellman g k=1m}; denote the corresponding optimal solution of \eqref{eq:bellman g k=1m} by $\bm{z}_{k}(d)$.
            \State $\bm{z}_{[k]} (d) \leftarrow (\bm{z}_{[k-1]}(d- \delta_{k} \indicate{\bm{z}_k (d) \neq \bm{z}^i_k}), \bm{z}_{k} (d))$
        \EndFor 
        \State \textbf{Solve} $g^i(m+1, 0)$ by Bellman equation \eqref{eq:bellman g k=m+1}; denote the corresponding optimal solution of \eqref{eq:bellman g k=m+1} by $d^\star$. 
        \Statex \textbf{return $g^i(m+1,0)$ and $\bm{z}_{[m]}(d^\star)$} 
    \end{algorithmic}
\end{algorithm}

\subsection{Graph-based Reformulation} \label{sec:graph}

Although the cutting-plane method does scale better than solving problem \eqref{eq:convex reformulation} monolithically, it is still limiting in the sizes of datasets that it can handle, as we will show later in Section~\ref{sec:runtime}. This limitation motivates us to propose a new framework for solving problem~\eqref{eq:convex reformulation}.

Our idea is to convert each of the constraints indexed by $(i,\bm{z}) \in [N] \times \mathcal{C}$ in problem \eqref{eq:convex reformulation} into a more tractable form, resulting in a formulation that can be solved directly. For each data point indexed by $i \in [N]$, 
we recall the original constraint set in problem \eqref{eq:convex reformulation},
$$ r_i \geq l_{\bm{\beta}} (\bm{x}^i, \bm{z}, y^i) - \lambda \sum_{\ell \in [m]} \delta_{\ell} \indicate{\bm{z}_{\ell} \neq \bm{z}^i_{\ell}}, \; \forall \bm{z} \in \mathcal{C}$$ 

By applying proper exponentiation operations, we define an equivalent optimization problem over all possible realizations of the categorical features $\mathcal{C}$ and establish an inequality that $y^i \left( \bm{\beta}_{\mathrm{x}}^\top \bm{x}^i + \beta_0 \right) $ is greater than or equal to:
\begin{equation}\label{eq:constraint reformulation}
\begin{aligned}
\max_{\bm{z} \in \mathcal{C}}  & \;\; -y^i \bm{\beta}_{\mathrm{z}}^\top \bm{z} - \log \Big( -1 \\
& \quad +\exp \Big( r_i + \lambda \sum_{\ell=1}^{m} \delta_{\ell} \indicate{\bm{z}_{\ell} \neq \bm{z}^i_{\ell}} \Big) \Big).
\end{aligned}
\end{equation}
This inequality holds if and only if the constraints for all $\bm{z} \in \mathcal{C}$ are satisfied. 
We analyze problem \eqref{eq:constraint reformulation} using dynamic programming.
For the data point indexed by $i$, we choose the same state space $\mathcal{S}^i$.
We denote the optimal objective value of each subproblem by \( h^i: \mathcal{S}^i \rightarrow \mathbb{R}\). We set $h^i(0,0) :=0$ and denote by $h^i(m+1,0)$ the optimal objective value of problem \eqref{eq:constraint reformulation}. Due to its structural similarity to problem~\eqref{eq:identification}, we continue to set subproblem \eqref{eq:subproblem k=1m} for each state \( (k,d) \in \mathcal{S}^i_1 \) and define $h^i(k,d) := g^i(k,d)$ for all $(k,d) \in \mathcal{S}^i_1$.
Problem~\eqref{eq:constraint reformulation} can be solved recursively using the following Bellman equations.

If $k \in [m]$, similar to equation \eqref{eq:bellman g k=1m},
\begin{equation}\label{eq:bellman h k=1m}
\begin{array}{cl}
    & h^i(k,d) \; = \;  \underset{\bm{z}_{k} \in \mathcal{F}_{kd}}{\text{max}}  -y^i \bm{\beta}_{\mathrm{z}k}^\top \bm{z}_{k} + h^i\left(k-1, \right. \\
   & \qquad \qquad \qquad \qquad \qquad \left. d - \delta_{k} \indicate{\bm{z}_{k} \neq \bm{z}^i_{k}}\right).
\end{array}
\end{equation}
If $k = m+1$, 
\begin{equation}\label{eq:bellman h k=m+1}
\begin{aligned}
 h^i(m+1, 0)  \; = \; & \underset{d}{\text{max}} \; h^i(m, d) \\
    & \qquad - \log\left( -1 + \exp(r_i + \lambda d) \right) \\
    & \text{s.t.} \quad (m, d) \in \mathcal{S}^i_1.
\end{aligned}
\end{equation}

Note that problem \eqref{eq:bellman h k=m+1} differs from problem \eqref{eq:bellman g k=m+1} due to the differences between problem \eqref{eq:identification} and problem \eqref{eq:constraint reformulation}. 
We need to guarantee $ y^i \left( \bm{\beta}_{\mathrm{x}}^\top \bm{x}^i + \beta_0 \right)  \geq h^i(m+1, 0)$. Without directly solving this dynamic programming problem based on relaxed solutions, we aim to reformulate this dynamic program to replace the constraint set in problem~\eqref{eq:convex reformulation} with an equivalent longest path problem.  

\begin{figure}
    \centering
\includegraphics[width=1.0\linewidth]{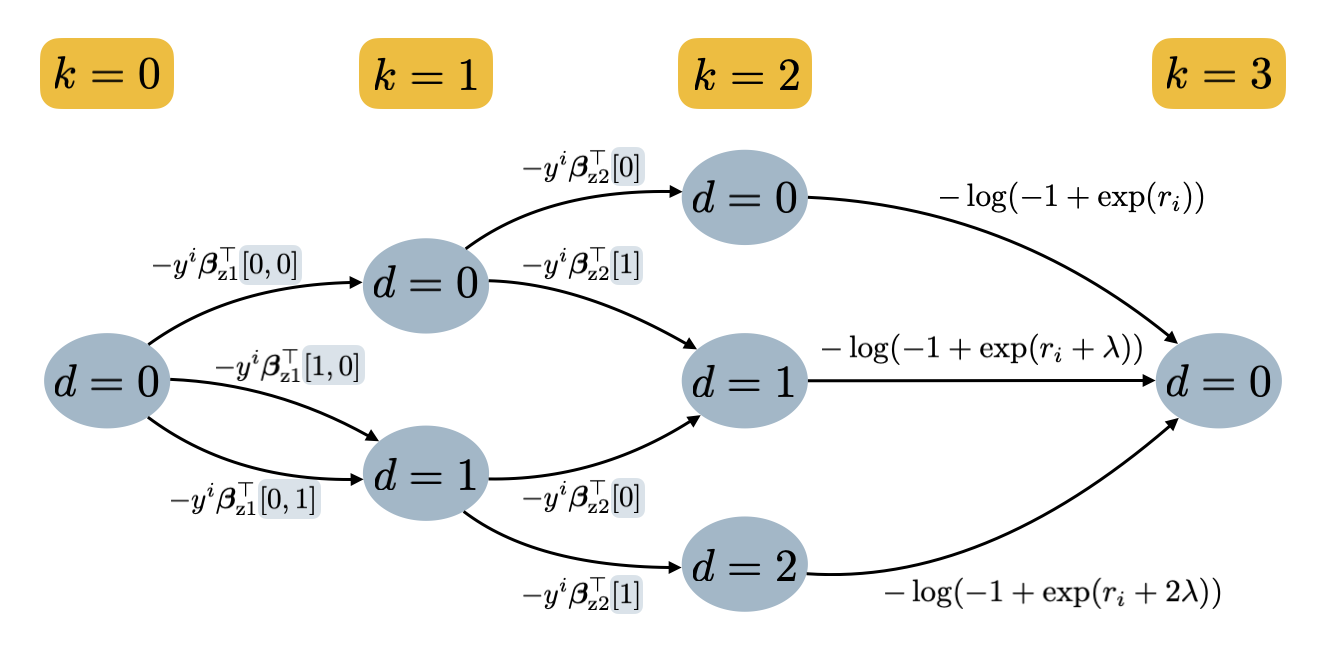}
    \caption{A graph example with only two categorical features processed by the one-hot encoding: $\bm{z} = (\bm{z}_1, \bm{z}_2)$ where $\bm{z}_1$ has three possible realizations: $[1,0], [0,1], [0,0]$ and  $\bm{z}_2$ has two possible realizations: $[1], [0]$.  
Given a data point with index $i$: $\bm{z}^i = (\bm{z}_1^i, \bm{z}_2^i) = ([0,0], [0])$. The weight parameters are $\bm{\delta} = (\delta_{1}, \delta_{2}) = (1,1)$.}
\label{fig:network_example}
\end{figure}

We define a weighted directed acyclic graph $\mathcal{G}^{i}=(\mathcal{V}^i, \mathcal{A}^i)$ to be the state transition graph of the dynamic program defined by equations \eqref{eq:bellman h k=1m}-\eqref{eq:bellman h k=m+1}, as illustrated in Figure~\ref{fig:network_example}. Formally, vertex set $\mathcal{V}^i$ corresponds to the set of states $\mathcal{S}^i$  and arc set $\mathcal{A}^i$ corresponds to the set of all possible transitions between the states. Each state $(k,d) \in \mathcal{S}^i$ has an associated vertex in $\mathcal{V}^i$. 
To express the arcs between vertices, for each $(k,d) \in \mathcal{S}^i$, 
\begin{enumerate}
\item If $k \in [m]$, 
for all $\bm{z}_k \in \mathcal{F}_{kd}$, 
create an arc from $(k-1,d- \delta_{k} \indicate{\bm{z}_{k} \neq \bm{z}^i_{k}})$ to 
$(k, d)$ with a weight $-y^i \bm{\beta}_{\mathrm{z}k}^\top \bm{z}_{k}$,

\item If $k=m+1$, for each $(m,d') \in \mathcal{S}^i$, create an arc from $(m,d')$ to $(m+1,0)$ with a weight $- \log\left( -1 + \exp(r_i + \lambda d') \right)$
\end{enumerate}

 We denote the arc weight function as $w^i: \mathcal{A}^i \rightarrow \mathbb{R}$. We also define the functions $s^i: \mathcal{A}^i \rightarrow \mathcal{V}^i$ and $t^i: \mathcal{A}^i \rightarrow \mathcal{V}^i$, which map each arc to its source and target vertices, respectively. 
 Based on the connection between the dynamic programming problem and graph structure, as well as the connection between the dynamic programming problem and problem \eqref{eq:constraint reformulation}, we present the following lemma.
 \begin{lem}\label{dp to path}
In each graph $\mathcal{G}^i$, the longest path from source $(0,0)$ to sink $(m+1,0)$ corresponds to the optimal solution of problem \eqref{eq:constraint reformulation} and the sum of the weights of this path equals its optimal objective value.
\end{lem}

The proof of Lemma \ref{dp to path} is provided in Appendix~\ref{sec:Proof of Lemma dp to path}. Using the connection in Lemma \ref{dp to path}, we can reformulate problem \eqref{eq:constraint reformulation} as a linear program that finds the longest path from the source $(0,0)$ to the sink $(m+1,0)$ in $\mathcal{G}^i$. We can generate a graph-based reformulation.
\begin{theorem}[Graph-based Formulation] \label{thm: graph-based reformulation}
    Problem \eqref{eq:convex reformulation} is equivalent to: 
\begin{equation}\label{eq:new_network_reformulation1}
\begin{aligned}
& \underset{\lambda,\boldsymbol{r},\bm{\beta}, \bm{\mu}}{\text{\emph{min}}}
& & \lambda \epsilon + \frac{1}{N}\sum_{i \in [N]} r_i \\
& \text{\emph{ \; s.t.}}
& &    y^i(\bm{\beta}_{\mathrm{x}}^\top \bm{x}^i +\beta_0)\geq -\mu_{(0,0)}^i + \mu_{(m+1,0)}^i, \; \forall i \in [N]  \\
& & &  \mu_{t^i(e)}^i - \mu_
{s^i(e)}^i \geq w^i(e), \; \forall i \in [N], e \in \mathcal{A}^i\\
& & & |\gamma_{j}^{-1} \beta_{\mathrm{x}j}| \leq \lambda, \; \forall j \in [n] \\
& & & \lambda \geq 0,\boldsymbol{r} \in \mathbb{R}^N,{\bm \beta} \in \mathbb{R}^{1+n+c}, \bm{\mu} \in \mathbb{R}^{\sum_{i \in [N]} |\mathcal{V}^i|}
\end{aligned}
\end{equation}
where $\bm{\mu} = (\bm{\mu}^1, \dots, \bm{\mu}^N)$ are dual variables of the longest path problems and the first two sets of constraints correspond to the dual formulation. 
\end{theorem}

The proof of Theorem \ref{thm: graph-based reformulation} is given in Appendix \ref{sec: Proof of Theorem graph-based formulation}. Compared to problem \eqref{eq:convex reformulation}, the size of this new formulation depends on the size of the dynamic programming digraphs. By selecting appropriate weight parameters~$\delta_{\ell}$, multiple arcs from different vertices can converge at the same vertex, significantly reducing the number of arcs and vertices in the graph. For example, rounding the weight parameters to some decimal places can further reduce the size of the digraphs because it can reduce the number of unique weighted distances. 
Additionally, most constraints in this reformulation are linear, except for those representing arcs targeting sink vertices, which are also convex and compatible with off-the-shelf solvers.

Unlike cutting-plane methods, the graph-based formulation can be solved in polynomial time with respect to the digraph size, as there are $N$ graphs for $N$ data points. Additionally, it is independent of the number of iterations, which is often difficult to control in cutting-plane methods.

\section{NUMERICAL RESULTS} 
\begin{table*}[ht]
\footnotesize
        \caption{Mean runtime in seconds of the graph-based formulation (Graph), the cutting-plane method with the most violated constraint identified by dynamic programming (DP), and the cutting-plane method proposed by \citet{closest} (Cutting), under three settings: all weight parameters set to 1 (Weights = 1), calibrated weight parameters rounded to integers (Integer Weights), and calibrated weight parameters rounded to one decimal place (1 Decimal Weights). 
         NaN in this table means that there is no numerical feature in the corresponding dataset. Following the notation in Section \ref{sec:notation and problem def}, $N$ is the number of data points, $n$ is the number of numerical features, $m$ is the number of categorical features, and $c$ is the number of categorical features after applying one-hot encoding. } 
    \centering
\begin{tabular}{|l|llll|lll|ll|ll|}
    \hline
\multicolumn{5}{|l|}{\textbf{}} & \multicolumn{3}{|l|}{Weights = 1} & \multicolumn{2}{|l|}{Integer Weights} & \multicolumn{2}{|l|}{1 Decimal Weights} \\ \hline

        Dataset & $N$ & $n$ & $m$ & $c$ & Graph & DP & Cutting & Graph & DP & Graph & DP \\ \hline    
         audiology & 226 & NaN & 62 & 85 & 82.66 & 3880.84 & 825.73 & 142.24 & 4461.94 & 5391.81 & 45345.44 \\ \hline
        balance-scale & 576 & NaN & 4 & 16 & 1.27 & 33.68 & 34.66 & 2.18 & 32.61 & 2.84 & 35.37 \\ \hline
        breast-cancer & 286 & NaN & 9 & 32 & 2.43 & 28.61 & 16.02 & 5.63 & 43.81 & 30.84 & 142.66 \\ \hline
        car & 1728 & NaN & 6 & 15 & 5.35 & 314.00 & 206.88 & 10.45 & 384.24 & 31.51 & 640.59 \\ \hline
        hayes-roth & 132 & NaN & 3 & 9 & 0.11 & 1.69 & 1.55 & 0.16 & 1.74 & 0.18 & 1.77 \\ \hline
        tic-tac-toe & 958 & NaN & 9 & 18 & 6.01 & 161.15 & 85.83 & 10.23 & 203.15 & 83.08 & 789.70 \\ \hline
        spect & 267 & NaN & 22 & 22 & 7.76 & 47.36 & 12.66  & 9.49 & 51.72 & 141.57 & 473.47 \\ \hline
        voting & 435 & NaN & 16 & 32 & 8.50 & 112.72 & 27.59 & 16.24 & 177.11 & 223.72 & 1253.95 \\ \hline
        credit-approval & 690 & 6 & 9 & 36 & 6.63 & 193.02 & 69.64 & 13.94 & 256.21 & 77.12 & 704.77 \\ \hline
         cylinder & 539 & 19 & 14 & 43 & 14.32 & 399.14 & 77.39 & 32.08 & 511.33 & 439.47 & 3557.80 \\ \hline
        hepatitis & 155 & 6 & 13 & 23 & 1.62 & 15.41 & 6.09  & 3.52  & 21.83 & 37.50 & 120.09 \\ \hline
        nursery & 12960 & NaN & 8 & 26 & 97.25 & 23874.78 & 11360.82  & 225.05  & 33132.95 & 1593.70 & 67219.20 \\ \hline
        online-shopper & 12330 & 14 & 3 & 14 & 20.57 & 8685.08 & 8394.99  & 23.96  & 8919.10 & 33.48 & 9114.63 \\ \hline
    \end{tabular}
    \label{tab:runtime}
\end{table*}
We evaluate our methods on 13 UCI datasets with numerical and/or categorical features \citep{UCI}. The datasets vary in size, with the number of data points ranging from 132 to 12960, and the number of categorical features ranging from 3 to 62. All algorithms were implemented in Julia 1.8.5 \citep{Julia-2017} using the JuMP mathematical modeling language~\citep{jump} and executed on AMD epyc-7513 CPU in single-core mode. We use MOSEK 10 to solve all optimization problems~\citep{mosek}.
The code can be found at: \url{ https://github.com/QingshiSun/Robust_Logistic_Regression}.

\subsection{Runtime Comparison} \label{sec:runtime}
We focus on runtime comparisons between the cutting-plane method proposed by \citet{closest}, the cutting-plane method described in Section \ref{sec:cutting method}, and the direct solution of our proposed graph-based formulation as outlined in Section \ref{sec:graph}.

In our experiments, each model instance consists of a dataset, the ambiguity set parameters \( \gamma_j \), \( \delta_{\ell} \), and \( \epsilon \), as well as the rounding precision applied to the weight parameters.
Following the ambiguity set calibration procedure outlined in Section~\ref{sec:calibration}, for all \( j \in [n] \) and \( \ell \in [m] \), we determine \( \gamma_j \) and \( \delta_{\ell} \) by sampling the probabilities of certainty \( \rho_{\mathrm{x}j} \)  and \( \rho_{\mathrm{z}\ell} \) from the same normal distribution in each model instance. Specifically, we consider normal distributions with means of \( 0.6, 0.7, 0.8, \) and \( 0.9 \), each with a standard deviation of \( 0.2 \).
  For each numerical feature \( j \), the interval \( [l_j, u_j] \) is set to \([-0.4\sigma_j, 0.4\sigma_j]\), where \( \sigma_j \) is the standard deviation of \( x_j \) in the training dataset. We also vary the level of robustness \( \theta \) by selecting a value from \{0.5, 0.65, 0.75, 0.8, 0.85, 0.9, 0.93, 0.96, 0.99\} for each model instance, which is then used to set the ambiguity set radius.  As discussed in Section \ref{sec:graph}, the sizes of the dynamic programming problems and the corresponding graph-based formulation depend on the chosen weight parameters. We evaluate three rounding precision cases for the weight parameters: (i) all weights are set to one, disregarding differences in the distribution shift likelihood; (ii) calibrated weight parameters are rounded to integer; and (iii) calibrated weight parameters are rounded to one decimal place. In total, for the former case, we evaluate 117 instances, with 9 instances per dataset, while for each of the latter two cases, we evaluate 468 model instances, with 36 instances per dataset.

The average runtime results are presented in Table~\ref{tab:runtime}. From the table, it can be seen that our graph-based formulation has significant improvement in runtime in all instances. Specifically, the graph-based formulation can be solved up to 408.12 times faster on average compared to the cutting-plane method proposed by \citet{closest}.

\subsection{Performance under Distribution Shifts} \label{sec:performance under shifts}

To evaluate the robustness of our proposed model under distribution shifts, we generate 5,000 perturbed test sets for each instance. Each perturbed test set is created by independently perturbing the original test data based on expected perturbations. Specifically, for a given set of sampled probabilities of certainty \( \rho_{\mathrm{x}j} \) and \( \rho_{\mathrm{z}\ell} \), we generate the perturbed test sets using the probability density function \eqref{eq:numerical pdf} for numerical features and the probability mass function \eqref{eq:categorical pmf} for categorical features.

Additionally, we assess the robustness of our method against \emph{unexpected} distribution shifts. We repeat the process of generating perturbed test sets for each instance but introduce unexpected perturbations by altering the values of \( \rho_{\mathrm{x}j} \) and \( \rho_{\mathrm{z}\ell} \) used in our model. This evaluation is motivated by the fact that domain knowledge is sometimes uncertain and not precisely estimated.  
We test seven unexpected scenarios. We shift each $\rho_{\mathrm{x}j}$ and $\rho_{\mathrm{z}\ell}$ value down by 0.2 and perturb the test data in 5,000 different ways based on these adjusted probability values. The same procedure is repeated with $\rho_{\mathrm{x}j}$ and $\rho_{\mathrm{z}\ell}$ shifted down by 0.1 and up by 0.1. In addition, we uniformly sample new $\rho_{\mathrm{x}j}$ and $\rho_{\mathrm{z}\ell}$ values for each feature within a neighborhood of radius 0.05 around the original $\rho_{\mathrm{x}j}$ and $\rho_{\mathrm{z}\ell}$ respectively, and perturb the test data in 5,000 different ways using these new values. This procedure is also applied for neighborhood radii of 0.1, 0.15, and 0.2. We denote the modified probabilities used to generate unexpectedly perturbed test sets as $\tilde{\rho}_{\mathrm{x}j}$ and $\tilde{\rho}_{\mathrm{z}\ell}$.

We evaluate the performance of logistic regression using two common metrics: adaptive calibration error \citep{calibrationerror} and AUC. These metrics are essential in high-stakes applications where predicted probabilities and their rankings, rather than just the final classifications, are of greater importance. 
For each model instance, we evaluate worst-case performance under expected perturbations by identifying the highest calibration error or lowest AUC across the 5,000 corresponding perturbed test sets for a given set of probability of certainty values \( \rho_{\mathrm{x}j} \) and \( \rho_{\mathrm{z}\ell} \). The average performance is measured by computing the mean over 5000 perturbed test sets. Similarly, we assess worst-case and average performance under unexpected perturbations using the corresponding 5,000 perturbed test sets generated with \( \tilde{\rho}_{\mathrm{x}j} \) and \( \tilde{\rho}_{\mathrm{z}\ell} \).

\begin{figure}[t!]
  \centering
  {\small
  \begin{subfigure}{0.495\columnwidth}
    \centering
    \captionsetup{justification=centering}
    \includegraphics[width=\linewidth]{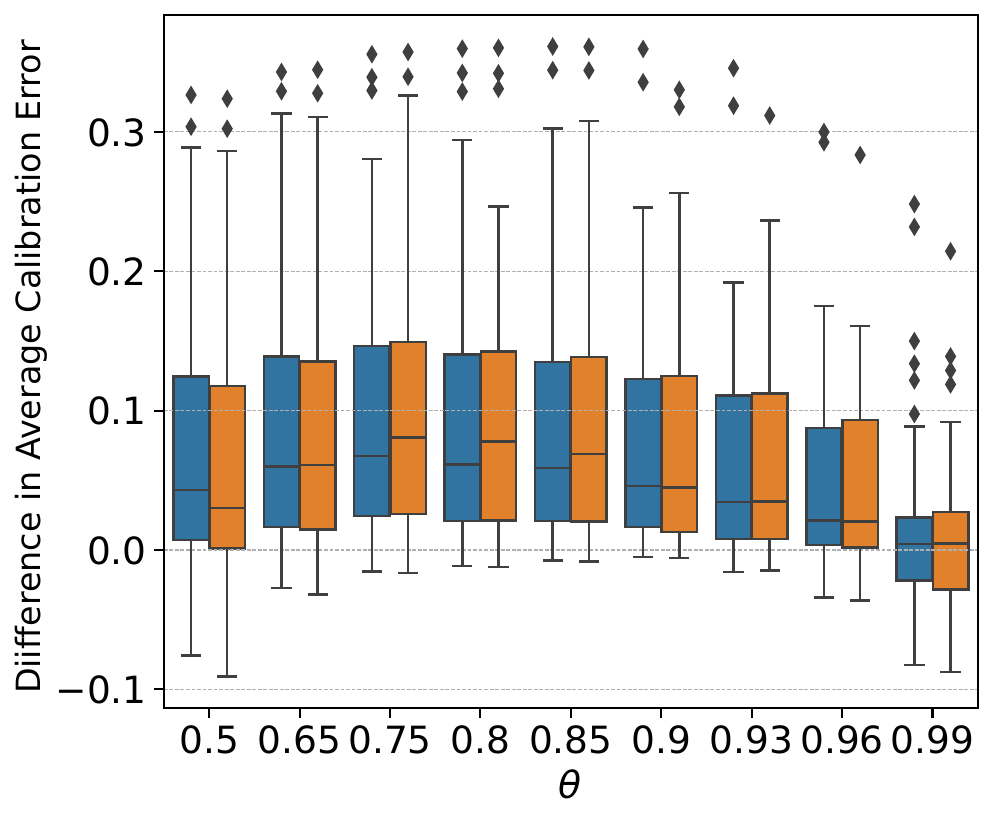}
  \end{subfigure}%
  \hfill
  \begin{subfigure}{0.495\columnwidth}
    \centering
    \captionsetup{justification=centering} 
    \includegraphics[width=\linewidth]{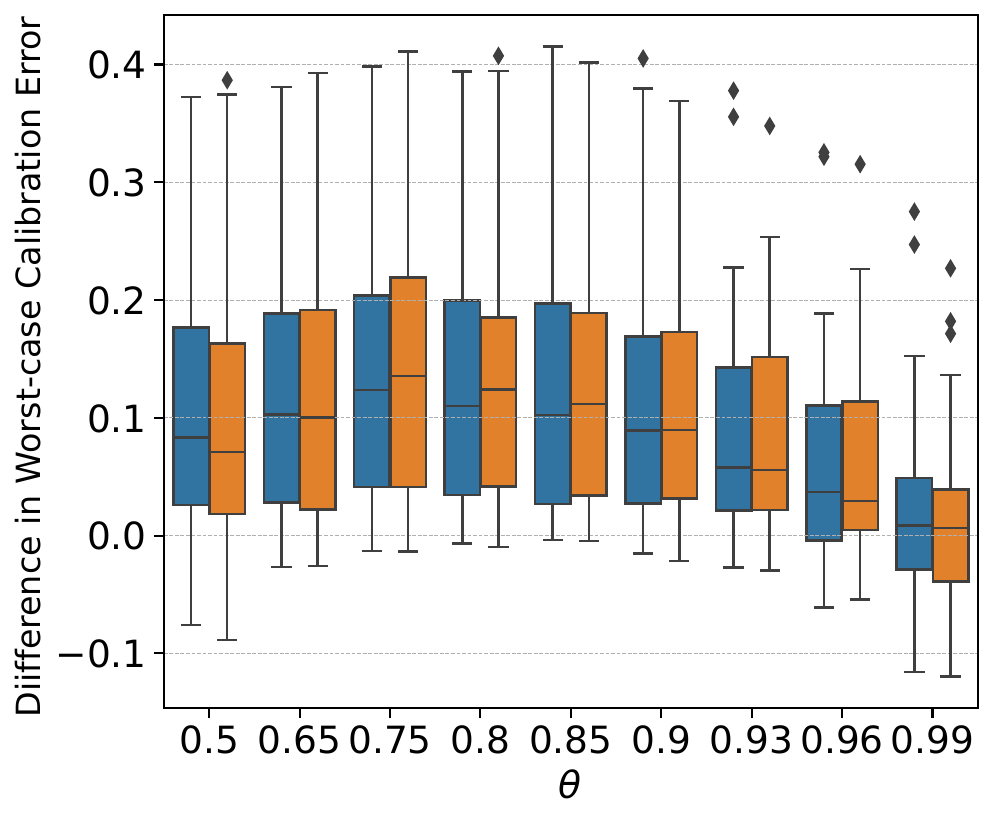}
  \end{subfigure}
  
  \vspace{-0.05cm}
  
  \begin{subfigure}{0.495\columnwidth}
    \centering
    \captionsetup{justification=centering}
    \includegraphics[width=\linewidth]{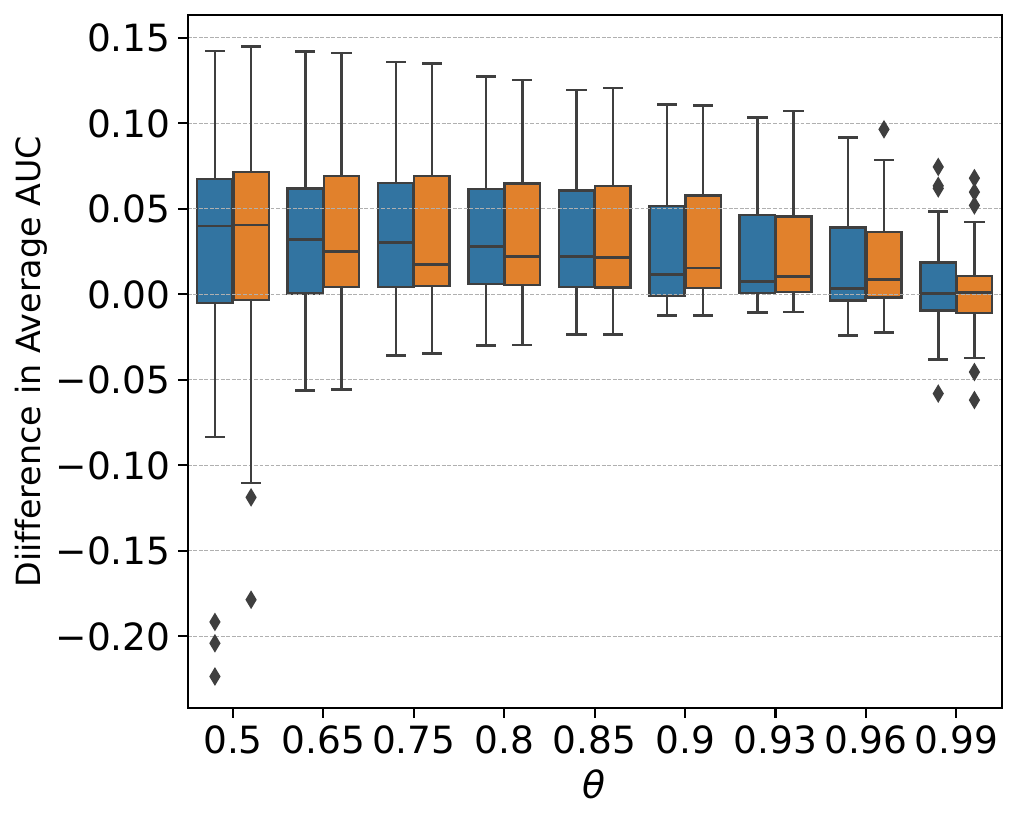}
  \end{subfigure}%
  \hfill
  \begin{subfigure}{0.495\columnwidth}
    \centering
    \captionsetup{justification=centering}
    \includegraphics[width=\linewidth]{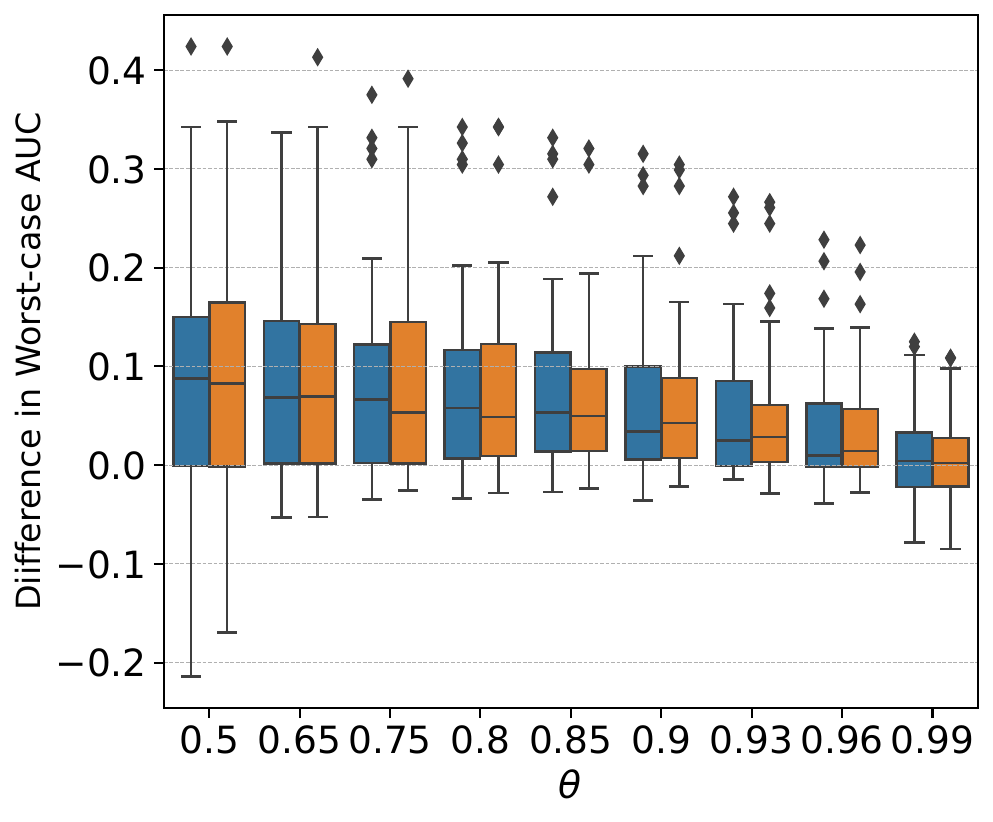}
  \end{subfigure}
  }
  \caption{\small Performance improvement compared to lasso logistic regression in terms of
calibration error and AUC across different levels of robustness $\theta$ under expected perturbations. Blue boxes: proposed models with weight parameters rounded to integer; orange boxes: with weight parameters rounded to one decimal.}
  \label{fig:merged_comparison_with_lasso}
\end{figure}

We compare our proposed models, trained with rounding precisions to the nearest integer and one decimal place, against two existing methods: lasso regularized logistic regression and distributionally robust logistic regression with all weight parameters set to 1, as studied by \citet{closest}.  The candidate lasso regularization coefficients are
$\{ 0, 5 \cdot 10^{-6}, 1 \cdot 10^{-5}, 5 \cdot 10^{-5}, 1 \cdot 10^{-4},  \ldots, 0.5, 1, 10, 100, 1000 \}$.
The ambiguity set radius candidates for the model proposed by \citet{closest} are:
$\{ 0, 10^{-5}, 10^{-4}, \ldots, 0.1, 1 \}$.
Both regularization coefficient and ambiguity set radius are selected using 5-fold cross-validation. Neither our proposed models nor the benchmarks models have access to the perturbed test set in the training phase.

\begin{figure}[t!]
   \centering
  {\small
  \begin{subfigure}{0.495\columnwidth}
    \centering
    \captionsetup{justification=centering}
    \includegraphics[width=\linewidth]{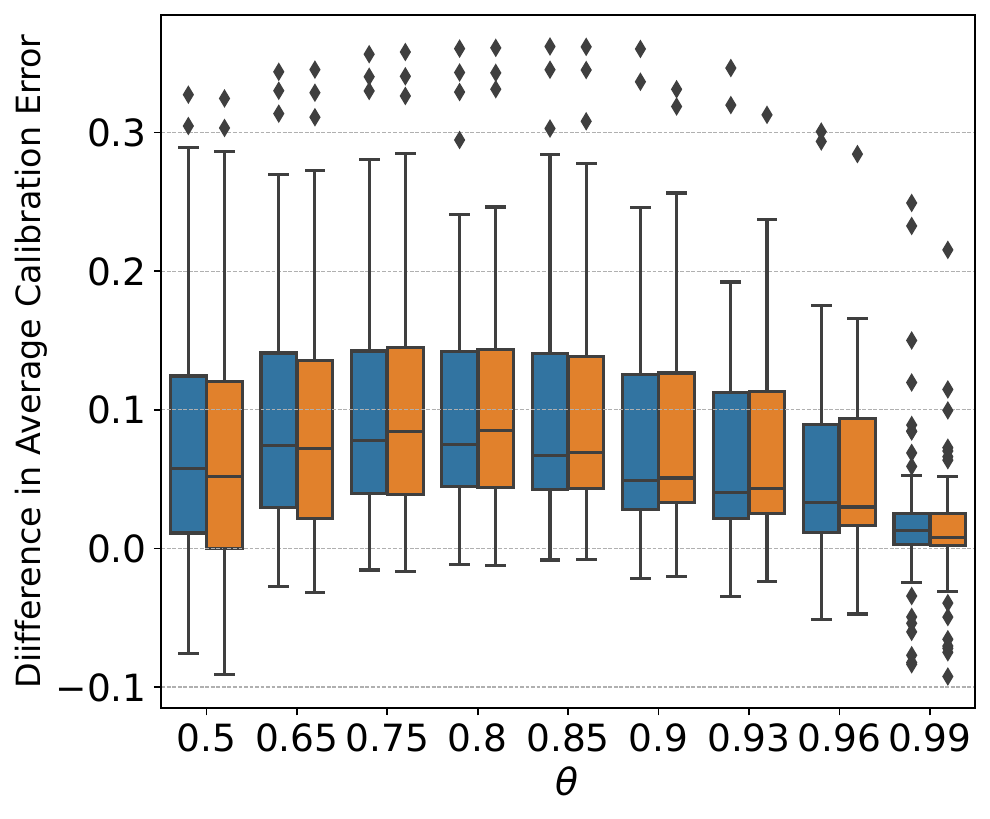}
  \end{subfigure}%
  \hfill
  \begin{subfigure}{0.495\columnwidth}
    \centering
    \captionsetup{justification=centering} 
    \includegraphics[width=\linewidth]{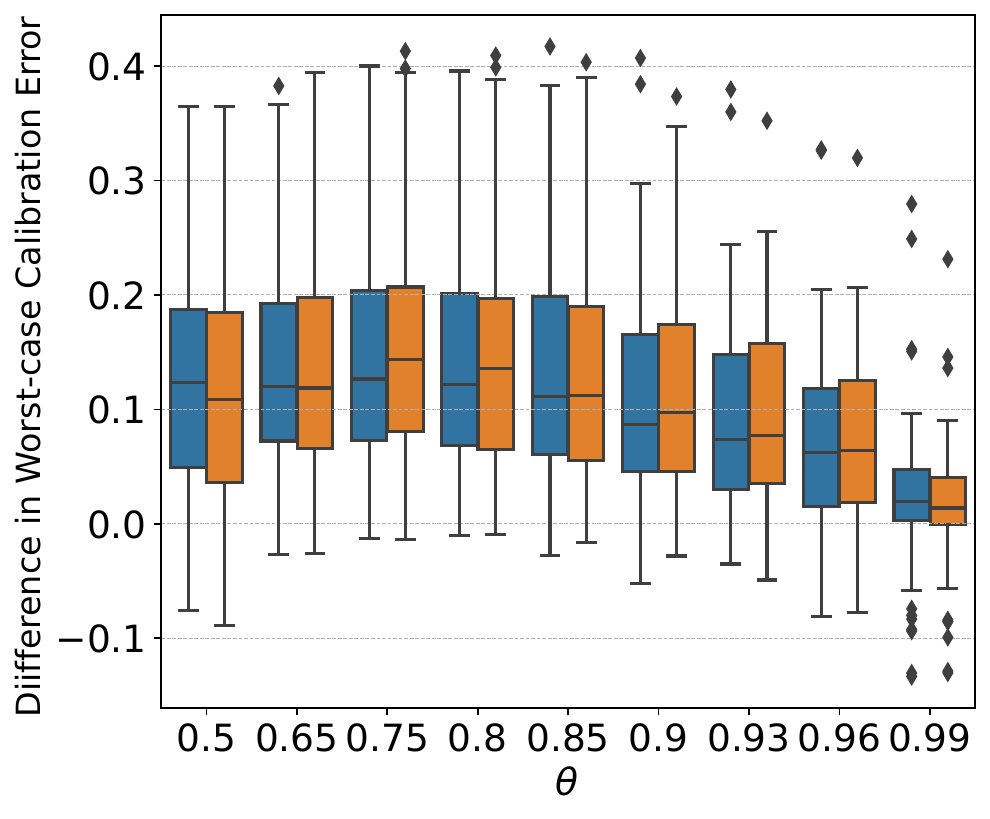}
  \end{subfigure}
  
  \vspace{-0.05cm}
  
  \begin{subfigure}{0.495\columnwidth}
    \centering
    \captionsetup{justification=centering}
    \includegraphics[width=\linewidth]{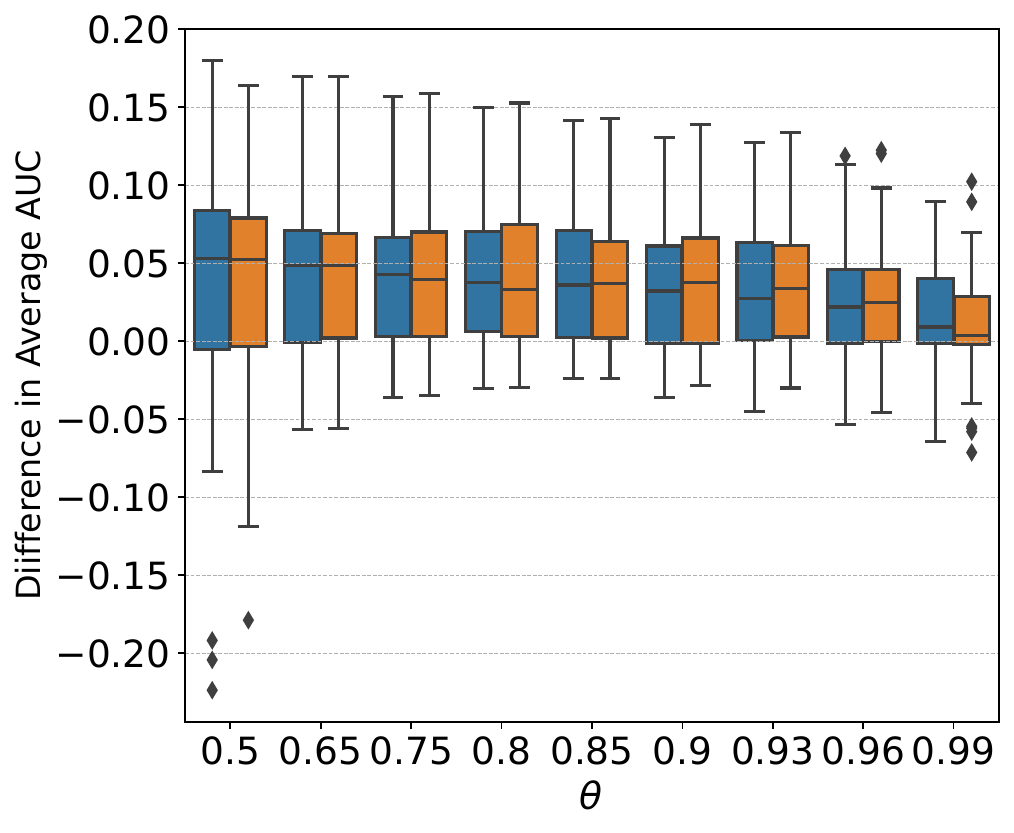}
  \end{subfigure}%
  \hfill
  \begin{subfigure}{0.495\columnwidth}
    \centering
    \captionsetup{justification=centering}
    \includegraphics[width=\linewidth]{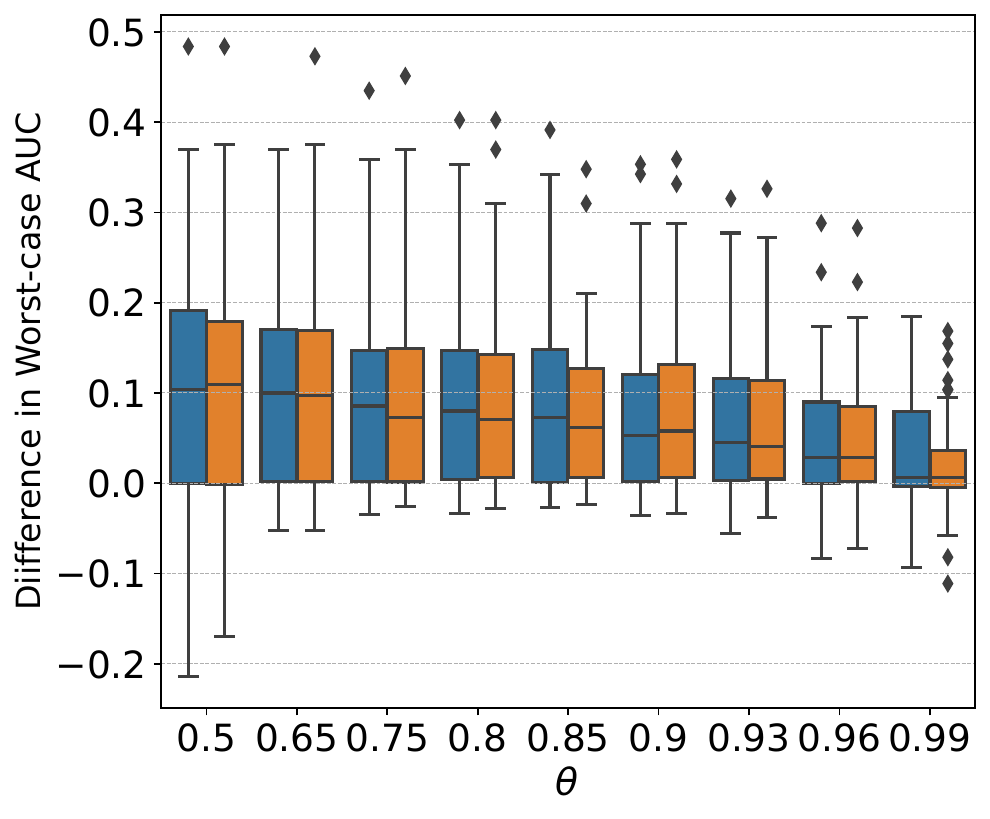}
  \end{subfigure}
  }
  \caption{\small Performance improvement compared to distributionally robust logistic regression with all weight parameters set to 1 in terms of
calibration error and AUC across different levels of robustness $\theta$ under expected perturbations. Blue boxes: proposed models with weight parameters rounded to integer; orange boxes: with weight parameters rounded to one decimal.}
  \label{fig:merged_comparison_with_m1}
\end{figure}

For each model instance, we compute the differences in worst-case and average-case performance between our proposed model and the benchmarks under identical expected and unexpected perturbations. A \emph{positive} difference indicates that our model outperforms the corresponding benchmark.
Figure \ref{fig:merged_comparison_with_lasso} summarizes the distribution of performance differences between our proposed model and lasso logistic regression across all tested instances under expected perturbations. Similarly, Figure \ref{fig:merged_comparison_with_m1} summarizes these differences relative to distributionally robust logistic regression with all weight parameters set to 1. Both figures demonstrate the consistent advantages of our model in terms of calibration error and AUC over a wide range of calibrated ambiguity set radii. 

We observe that a highly small ambiguity set radius (e.g., \( \theta = 0.99 \)) limits the model's effectiveness against distribution shifts, while an overly large radius results in excessive conservatism and degraded performance. Additionally, rounding weight parameters to integers, compared to one-decimal precision, maintains robust performance and significantly reduces runtime, enabling scalability to large-scale applications.

The distributions of worst-case and average-case improvements of our proposed model compared to benchmarks under unexpected perturbations are provided in Appendix \ref{sec:Unexpected Perturbations}.
These results show the robustness of our model in the presence of model misspecification.

\subsubsection*{Acknowledgments}
Phebe Vayanos and Qingshi Sun are funded in part by the National Science Foundation under CAREER award number 2046230. Nathan Justin is funded in part by the National Science Foundation Graduate Research Fellowship Program (GRFP).
Andr\'es G\'omez is funded in part by AFOSR grant No. FA9550-24-1-0086. The authors are grateful for the support and thank the anonymous reviewers for their detailed and insightful comments.

\bibliographystyle{plainnat}
\bibliography{main}

\appendix
\onecolumn
\aistatstitle{Supplementary Materials}

\section{DETAILS OF PARAMETER CALIBRATION}\label{sec:details of calibration}
We propose a calibration method for the parameters in the ambiguity set \eqref{eq:ambiguity set}, including the radius \( \epsilon \) and the weight parameters \( \gamma_{j} \) and \( \delta_{\ell} \), where \( j \in [n] \) and \( \ell \in [m] \). This method leverages the relationship between robust optimization (RO) and DRO. We first define a logistic regression model based on RO and establish a correspondence between the parameters in the uncertainty set of the robust logistic regression and those in the ambiguity set of our proposed distributionally robust logistic regression model. Subsequently, we extend the calibration method developed by \citet{decisiontree} for robust optimization from discrete to numerical features. The parameters in the ambiguity set are determined based on this correspondence and the calibrated parameters from the uncertainty set of robust optimization.

Compared to DRO, 
instead of constructing an ambiguity set containing a range of probability distributions and optimizing for the worst-case \emph{distribution} within that set, RO aims to protect against worst-case scenarios by considering the worst possible \emph{realization} of uncertain parameters within a given uncertainty set.  
 Theoretically,  \citet{GaoROvsDRO} proves that distributionally robust optimization problems can be approximated by robust optimization problems, and thereby some distributionally optimization problems can be processed by tools from robust optimization. 
In our setting, we define a logistic regression model based on robust optimization that accounts for the worst-case realization of perturbed features within an uncertainty set. For the training data, we
define perturbed numerical features ${\Tilde{X}} = [\bm{\Tilde{x}}^1, \bm{\Tilde{x}}^2,\dots,\bm{\Tilde{x}}^N]^\top$ and perturbed categorical features $\Tilde{Z} = [\bm{\Tilde{z}}^1, \bm{\Tilde{z}}^2,\dots,\bm{\Tilde{z}}^N]^\top$. We formulate a robust logistic regression problem without label perturbation
 \begin{equation}\label{eq:RO}
    \begin{array}{l l l}
        \displaystyle  \mathop{\text{minimize}}_{\bm{\beta} \in \mathbb{R}^{1+n+c}} \; \mathop{\text{maximize}}_{(\Tilde{X}, \Tilde{Z}) \in \mathcal{U}
        } & \displaystyle \dfrac{1}{N} \sum_{i \in [N]}  l_{\bm{\beta}} (\bm{\Tilde{x}}^i, \bm{\Tilde{z}}^i, y^i), 
    \end{array}
\end{equation}
where $\mathcal{U}$ is the uncertainty set defined as
\begin{equation} \label{eq:uncertainty set}
\begin{aligned}
    \mathcal{U} = \Bigl\{ & (\Tilde{X}, \Tilde{Z}) \mid \frac{1}{N} \sum_{i \in [N]} \Big( \sum_{j \in [n]} \gamma_{j}' | \Tilde{x}_j^i - x_j^i| + \sum_{\ell \in [m]} \delta_{\ell}' \indicate{\bm{\Tilde{z}}_{\ell}^i \neq \bm{z}_{\ell}^i} \Big) \leq \epsilon', \bm{\Tilde{x}}^i \in \mathbb{R}^n, \ \bm{\Tilde{z}}^i \in \mathcal{C}, \ i \in [N] \Bigr\},
\end{aligned}
\end{equation}
where $\epsilon'>0$ is the total allowable budget of uncertainty across data points, controlling the level of robustness for distribution shifts. $ \gamma_{j}'>0$ and $ \delta_{\ell}'>0$ are the weight parameters of the perturbation on numerical feature $j$ and on categorical feature $\ell$ respectively. $\gamma'_j$ is the cost of perturbing $x^i_j$ by one in either direction, and $\delta'_{\ell}$ is the cost of perturbing $\bm{z}^i_{\ell}$ to a different value in the set $\mathcal{C}_{\ell}$. Similar to our proposed distributionally robust logistic regression model, this robust logistic regression model also addresses sensitivity to feature shifts through weight parameters. In addition, although they differ in addressing distribution shifts—one focusing on worst-case realizations and the other on worst-case distributions—both models ultimately control the tolerance for distribution shifts using budget parameters.

We next build the specific connection between robust logistic regression problem \eqref{eq:RO} and our proposed distributionally robust logistic regression problem \eqref{eq:original formulation}. That is, problem \eqref{eq:original formulation} can be transformed  into the form of problem \eqref{eq:RO} by introducing appropriate constraints. 
By incorporating the uncertainty set $\mathcal{U}$, the inner problem of the robust logistic regression problem \eqref{eq:RO} is equivalent to
 \begin{equation}\label{eq:explicit RO}
    \begin{array}{l@{\quad}l@{\qquad}l}
        \displaystyle \mathop{\text{maximize}}_{\Tilde{X}, \Tilde{Z}} & \displaystyle \dfrac{1}{N} \sum_{i \in [N]}  l_{\bm{\beta}} (\bm{\Tilde{x}}^i, \bm{\Tilde{z}}^i, y^i )  \\[5mm]
         \displaystyle \text{subject to} 
        & \displaystyle \frac{1}{N} \sum_{i \in [N]} \left( \sum_{j \in [n]} \gamma_{j} | \Tilde{x}_j^i - x_j^i| + \sum_{\ell \in [m]} \delta_{\ell} \indicate{\bm{\Tilde{z}}_{\ell}^i \neq \bm{z}_{\ell}^i} \right)  \leq \epsilon'  \\[7mm]
        & (\bm{\Tilde{x}^i}, \bm{\Tilde{z}^i}) \in \mathbb{R}^n \times \mathcal{C}, \; i \in [N]. 
    \end{array}
\end{equation}
Following the proof of Theorem 1 by \citet{closest}, the inner problem of our proposed distributionally robust logistic regression \eqref{eq:original formulation} can be reformulated as 

\begin{equation*}
    \begin{array}{l@{\quad}l@{\qquad}l}
        \displaystyle \mathop{\text{maximize}}_{\mathbb{Q}^i} & \displaystyle \dfrac{1}{N} \sum_{i \in [N]} \int_{\bm{\xi} \in \Xi} l_{\bm{\beta}}(\bm{\xi}) \, \mathbb{Q}^i (\diff \bm{\xi}) \\[5mm]
        \displaystyle \text{subject to} & \displaystyle \dfrac{1}{N} \sum_{i \in [N]} \int_{\bm{\xi} \in \Xi} d(\bm{\xi}, \bm{\xi}^i) \, \mathbb{Q}^i (\diff \bm{\xi}) \leq \epsilon \\[5mm]
        & \displaystyle \mathbb{Q}^i \in \mathcal{P}_0(\Xi), \, i \in [N].
    \end{array}
\end{equation*}
where $\mathbb{Q}^i(\diff \bm{\xi}) := \Pi(\diff \bm{\xi} | \bm{\xi}^i)$ the conditional distribution of $\Pi$ upon the realization of $\bm{\xi}' = \bm{\xi}^i$.
We add proper constraints on the selection of conditional probabilities to obtain a new formulation
\begin{equation*} 
    \begin{array}{l@{\quad}l@{\qquad}l}
        \displaystyle \mathop{\text{maximize}}_{\mathbb{Q}^i} & \displaystyle \dfrac{1}{N} \sum_{i \in [N]} \int_{\bm{\xi} \in \Xi} l_{\bm{\beta}}(\bm{\xi}) \, \mathbb{Q}^i (\diff \bm{\xi}) \\[5mm]
        \displaystyle \text{subject to} & \displaystyle \dfrac{1}{N} \sum_{i \in [N]} \int_{\bm{\xi} \in \Xi} d(\bm{\xi}, \bm{\xi}^i) \, \mathbb{Q}^i (\diff \bm{\xi}) \leq \epsilon \\[5mm]
        & \displaystyle \mathbb{Q}^i \in \mathcal{P}_0(\Xi), \, i \in [N] \\[4mm]
        & \displaystyle \mathbb{Q}^i \in \left\{ \delta_{(\bm{\Tilde{x}^i}, \bm{\Tilde{z}^i}, y^i)} \mid (\bm{\Tilde{x}^i}, \bm{\Tilde{z}^i}) \in \mathbb{R}^n \times \mathcal{C} 
        , \, i \in [N] \right\},
    \end{array}
\end{equation*}

where $\delta_{(\bm{\Tilde{x}^i}, \bm{\Tilde{z}^i}, y^i)}$ is the Dirac delta function centered at $(\bm{\Tilde{x}^i}, \bm{\Tilde{z}^i}, y^i)$.  Expressing $d(\bm{\xi}, \bm{\xi}^i)$ in its explicit form, the above formulation is equivalent to 
\begin{equation} \label{eq:dro with additional constraints}
    \begin{array}{l@{\quad}l@{\qquad}l}
        \displaystyle \mathop{\text{maximize}}_{\Tilde{X}, \Tilde{Z}} & \displaystyle \dfrac{1}{N} \sum_{i \in [N]}  l_{\bm{\beta}} (\bm{\Tilde{x}}^i, \bm{\Tilde{z}}^i, y^i )  \\[5mm]
        \displaystyle \text{subject to} & \displaystyle \dfrac{1}{N} \sum_{i \in [N]} \left( \sum_{j \in [n]} \gamma_{j} | \Tilde{x}_j^i - x_j^i| + \sum_{\ell \in [m]} \delta_{\ell} \indicate{\bm{\Tilde{z}}_{\ell}^i \neq \bm{z}_{\ell}^i} \right)  \leq \epsilon \\ [7mm]
         & (\bm{\Tilde{x}^i}, \bm{\Tilde{z}^i}) \in \mathbb{R}^n \times \mathcal{C}, \; i \in [N]. 
    \end{array}
\end{equation}

Now, problem \eqref{eq:dro with additional constraints} is in the form of \eqref{eq:explicit RO}. 
 The correspondence between parameters in the ambiguity set of the distributionally robust logistic regression model and parameters in the uncertainty set of the robust logistic regression problem model is $\gamma_{j} = \gamma_{j}'$, $\delta_{j} = \delta_{j}'$, and $\epsilon = \epsilon'$. 

Next, we extend the calibration method for robust optimization problems studied by \citet{decisiontree} to the case with both numerical and categorical features. 
Based on the assumptions in Section \ref{sec:calibration}, we assume the perturbation on numerical feature $j$ follows a Laplace distribution $f$ with scale parameter $b_j$ and location parameter 0. 
Since the probability of certainty $\rho_{\mathrm{x}j}$ represents the probability that the perturbation on numerical feature $j$ falling between $l_j$ and $u_j$ during deployment, for each data point indexed by $i$,
we can express $b_j$ through
\begin{equation*}
    \mathbb{P}(l_j \leq  \Tilde{x}_j^i - x_j^i \leq u_j ) = \int_{-u_j + x_j^i}^{u_j + x_j^i} f(\Tilde{x}_j^i) \text{d}\Tilde{x}_j^i = \rho_{\mathrm{x}j}.
\end{equation*}
We obtain $b_j = \frac{-u_j}{\log(1-\rho_{\mathrm{x}j})}$.

We now build uncertainty sets using hypothesis testing \citep{datadrivenRO, decisiontree}.  We set up a likelihood ratio test on the magnitude of the perturbation with threshold $\theta^N$, where the exponent $N$ to normalize across different datasets with varying number of data points.  Our null hypothesis is that the perturbed numerical and categorical features, $\tilde{X}$ and $\tilde{Z}$ come from the distributions described in Section \ref{sec:calibration}. If this null hypothesis fails to be rejected, then $\tilde{X}$ and $\tilde{Z}$ lie within our uncertainty set. That is, $\tilde{X}$ and $\tilde{Z}$ lies within the uncertainty set if the following inequality 
\begin{equation}\label{ratio}
    \frac{
\prod_{i \in [N]} \left( 
    \prod_{j \in [n]} \left( \frac{1}{2b_j} \exp \left( -\frac{\lvert \tilde{x}_{j}^i - x_j^i \rvert}{b_j} \right) \right) 
    \cdot 
    \prod_{\ell \in [m]} \left( \indicate{\tilde{\bm{z}}_{\ell}^i = \bm{z}_{\ell}^i} \rho_{\mathrm{z}\ell} 
    +  \indicate{\tilde{\bm{z}}_{\ell}^i \neq \bm{z}_{\ell}^i} \frac{1 - \rho_{\mathrm{z}{\ell}}}{ |\mathcal{C}_{\ell}| - 1} \right) 
\right) 
}{
 \prod_{i \in [N]} \left( \prod_{j \in [n]} \frac{1}{2b_j} \cdot \prod_{\ell \in [m]} \rho_{\mathrm{z}\ell} \right)
}
\geq \theta^N
\end{equation}
is satisfied. 
The numerator of the left hand side of inequality \eqref{ratio} is the likelihood under the null hypothesis. The denominator of the left hand side is the likelihood of the most probable realization, meaning the likelihood of no perturbation.

By applying logarithm operations at both sides, we reduce inequality \eqref{ratio} to
\begin{equation*}
   \sum_{i \in [N]} \left( \sum_{j \in [n]} \frac{| \Tilde{x}_j^i - x_j^i|}{b_j} \right)  
   + \sum_{\ell \in [m]} \left( -\log \left( \indicate{\tilde{\bm{z}}_{\ell}^i = \bm{z}_{\ell}^i} 
   + \indicate{\tilde{\bm{z}}_{\ell} \neq \bm{z}_{\ell}^i} \frac{1 - \rho_{\mathrm{z}\ell}}{ \rho_{\mathrm{z}\ell} (|\mathcal{C}_{\ell}| - 1)} \right) \right)  
   \leq -N \log \theta,
\end{equation*}
For categorical features, since when no perturbation exists, the cost is 0. We rewrite it as
\begin{equation} \label{ineq: ratio final}
  \frac{1}{N} \left( \sum_{i \in [N]} \left( \sum_{j \in [n]} \frac{| \Tilde{x}_j^i - x_j^i|}{b_j} \right)  
   + \sum_{\ell \in [m]} \left( -\log \left( 
\frac{1 - \rho_{\mathrm{z}\ell}}{ \rho_{\mathrm{z}\ell} (|\mathcal{C}_{\ell}| - 1)} \right)\indicate{\tilde{\bm{z}}_{\ell} \neq \bm{z}_{\ell}^i} \right) \right) 
   \leq -\log \theta,
\end{equation}

Since \eqref{ineq: ratio final} is the form of uncertainty set \eqref{eq:uncertainty set}, we can set 
\begin{equation*}
     \delta_{\ell}' = \log \left( \frac{\rho_{\mathrm{z}\ell} (|\mathcal{C}_{\ell}| - 1)}{1 - \rho_{\mathrm{z}\ell}} \right),
\end{equation*}
\begin{equation*}
    \gamma_{j}' = \frac{-\log(1 - \rho_{\mathrm{x}j})}{u_j}
\end{equation*}
\begin{equation*}
    \epsilon' = - \log \theta
\end{equation*}

Based on the correspondence $\gamma_{j} = \gamma_{j}'$, $\delta_{j} = \delta_{j}'$, and $\epsilon = \epsilon'$, we set
\begin{equation*}
     \delta_{\ell} = \log \left( \frac{\rho_{\mathrm{z}\ell} (|\mathcal{C}_{\ell}| - 1)}{1 - \rho_{\mathrm{z}\ell}} \right),
\end{equation*}
\begin{equation*}
    \gamma_{j} = \frac{-\log(1 - \rho_{\mathrm{x}j})}{u_j}
\end{equation*}
\begin{equation*}
    \epsilon = - \log \theta
\end{equation*}

\section{CUTTING-PLANE SCHEME} \label{sec:details of cutting-plane scheme}
\begin{algorithm}[h]
    \begin{algorithmic}[1]
        \caption{Cutting-Plane Scheme for Problem~\eqref{eq:convex reformulation}.} \label{alg:cac_gen}
        \Statex \textbf{Input:} constraint set $\mathcal{W} \subseteq [N] \times \mathcal{C}$.
        \Statex \textbf{Output:} optimal solution $( \lambda^\star, \bm{\beta}^\star, \bm{r}^\star)$ to problem \eqref{eq:convex reformulation}
        \State \textbf{Initialize} $\text{LB}_0 = -\infty$ and $\text{UB}_0 = +\infty$ as lower and upper bounds for problem \eqref{eq:convex reformulation}; 
        \While{$\text{LB}_t < \text{UB}_t$}
            \State Let $e^\star$ be the optimal value and $( \lambda, \bm{\beta}, \bm{r})$ be the optimal solution to the relaxed version of~\eqref{eq:convex reformulation}, where the original constraint set $[N] \times \mathcal{C}$ is replaced by $\mathcal{W}$.
            \For{$i \in [N]$}
            \State Identify the most violated constraint by Algorithm \ref{alg:dp}; get violation $\vartheta_i$ and corresponding solution $\bm{z}(i)$.
            \EndFor
            \State Calculate $\vartheta^\star = \max_{i \in [N]} \{\vartheta_i\}$ and $i^\star = \text{argmax}_{i \in [N]} \{\vartheta_i\}$.
            \State Add the constraint indexed by $(i^\star, \bm{z}(i^\star))$ to $\mathcal{W}$.
            \State Update $\text{LB}_t = e^\star$ and $\text{UB}_t = \min \{ \text{UB}_{t-1}, \; e^\star + \vartheta^\star \}$
            \State Update $t = t + 1$
        \EndWhile
        \State \textbf{return} solution $( \lambda, \bm{\beta}, \bm{r})$.
    \end{algorithmic}
\end{algorithm}
We adapt the cutting-plane algorithm proposed by \citet{closest}, as described in Algorithm \ref{alg:cac_gen}. This algorithm can solve problem \eqref{eq:convex reformulation} in finite many iterations, following the proof of 
Theorem 4 by \citet{closest}.

\section{DETAILS OF DATA PROCESSING }
For each dataset, missing values in categorical features are treated as a separate category within that feature, while missing values in each numerical feature are replaced by the median of the remaining values of that feature. Features with only one unique category in the raw dataset are ignored. For datasets with non-binary labels, we convert the labels into binary by distinguishing the majority class from all other labels. Additionally, unrelated columns are removed based on dataset-specific properties. For example, the "name" and "hobby" columns in the "hayes-roth" dataset are deleted because they consist of randomly generated values. Additionally, we convert the text in certain columns to lowercase when the upper and lower case have the same meaning, to prevent the model from treating them as distinct categories.

\section{DISTRIBUTIONS OF PERFORMANCE IMPROVEMENTS UNDER UNEXPECTED PERTURBATIONS} \label{sec:Unexpected Perturbations}
\begin{figure*}[htbp]  
    \centering
    \begin{subfigure}[b]{0.49\linewidth}
        \centering
        \includegraphics[width=\linewidth]{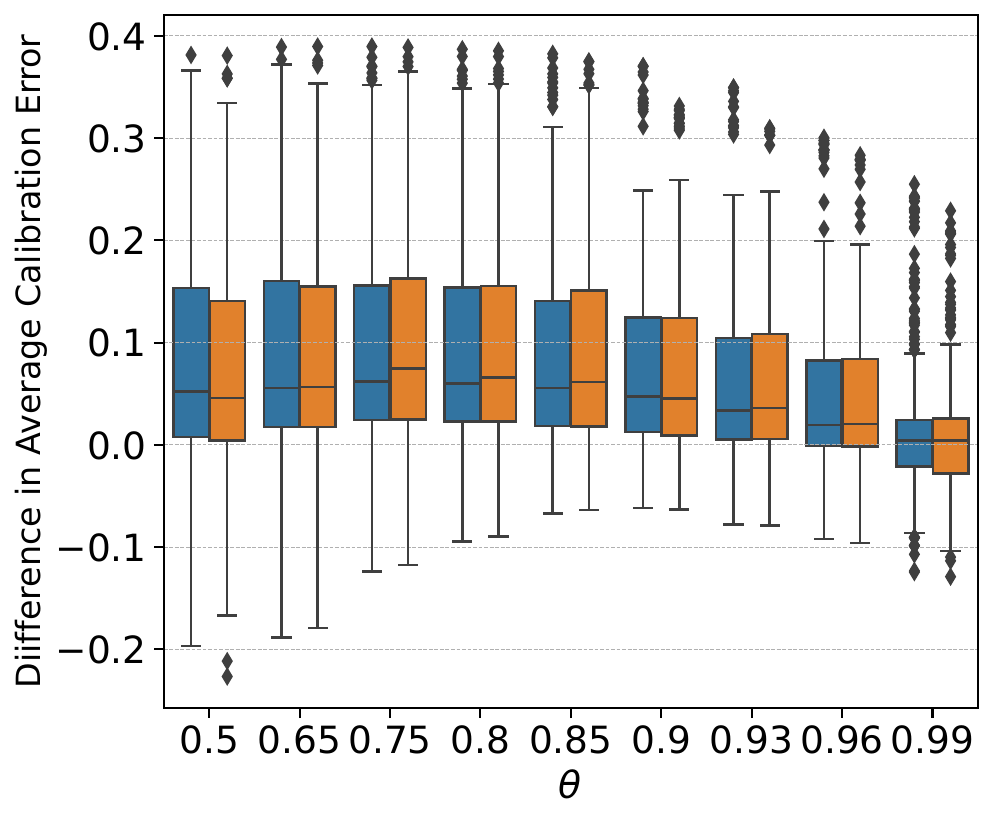}
    \end{subfigure}
    \hfill
    \begin{subfigure}[b]{0.49\linewidth}
        \centering
        \includegraphics[width=\linewidth]{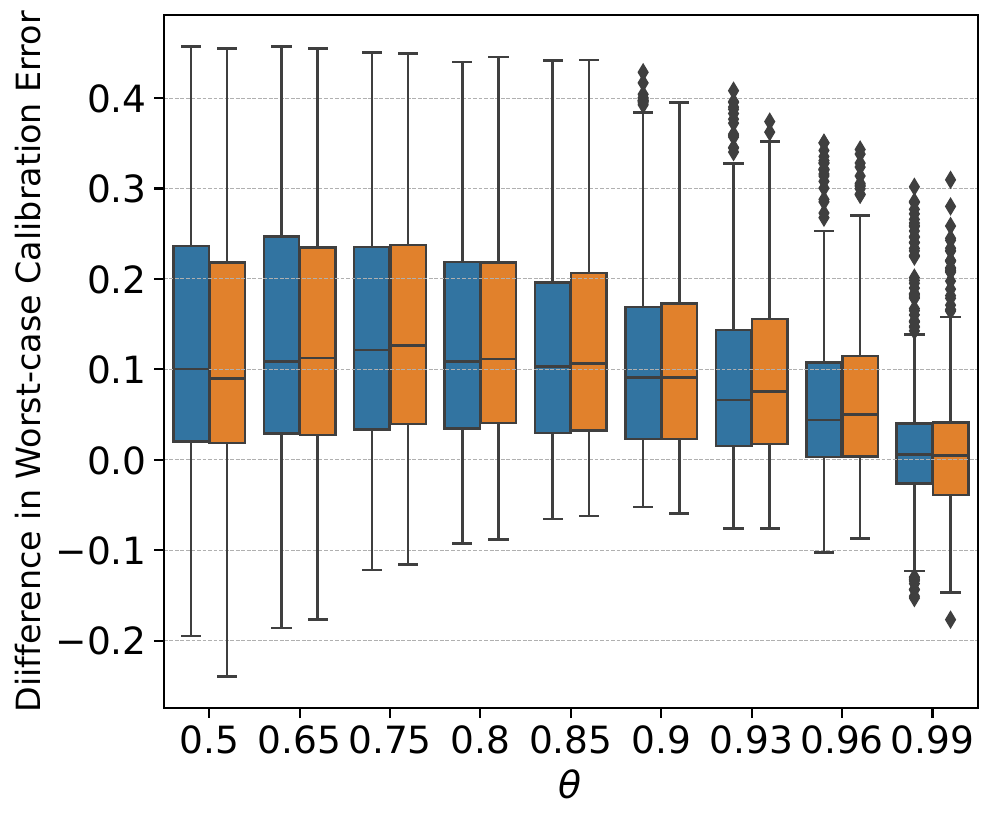}
    \end{subfigure}
    
    \vspace{0.2cm} 

    \begin{subfigure}[b]{0.49\linewidth}
        \centering
        \includegraphics[width=\linewidth]{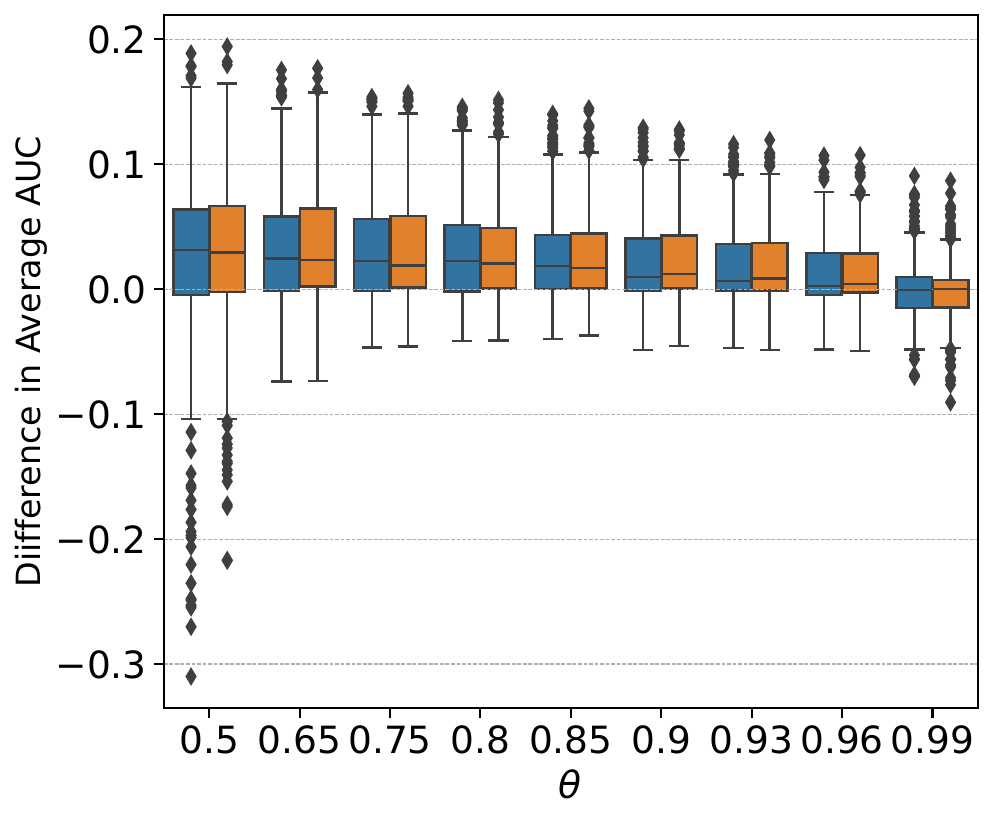}
    \end{subfigure}
    \hfill
    \begin{subfigure}[b]{0.49\linewidth}
        \centering
        \includegraphics[width=\linewidth]{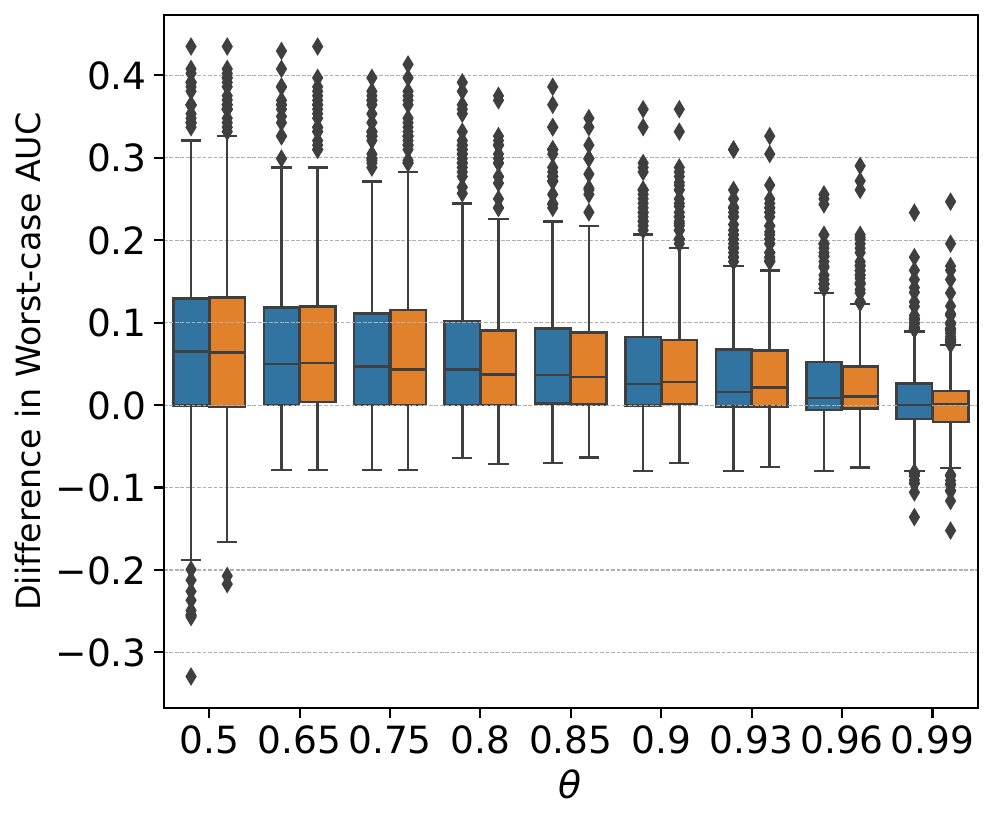}
    \end{subfigure}

    \caption{Overview of performance improvement compared to lasso-regularized logistic regression in terms of calibration error and AUC across different levels of robustness $\theta$ under \emph{unexpected} perturbations. The blue boxes represent our proposed models with calibrated parameters rounded to integer. The orange boxes represent  our proposed models with calibrated parameters rounded to one decimal place. }
    \label{fig:comparison with lasso un}
\end{figure*}

\begin{figure*}[htbp] 
    \centering
    \begin{subfigure}[b]{0.49\linewidth}
        \centering
        \includegraphics[width=\linewidth]{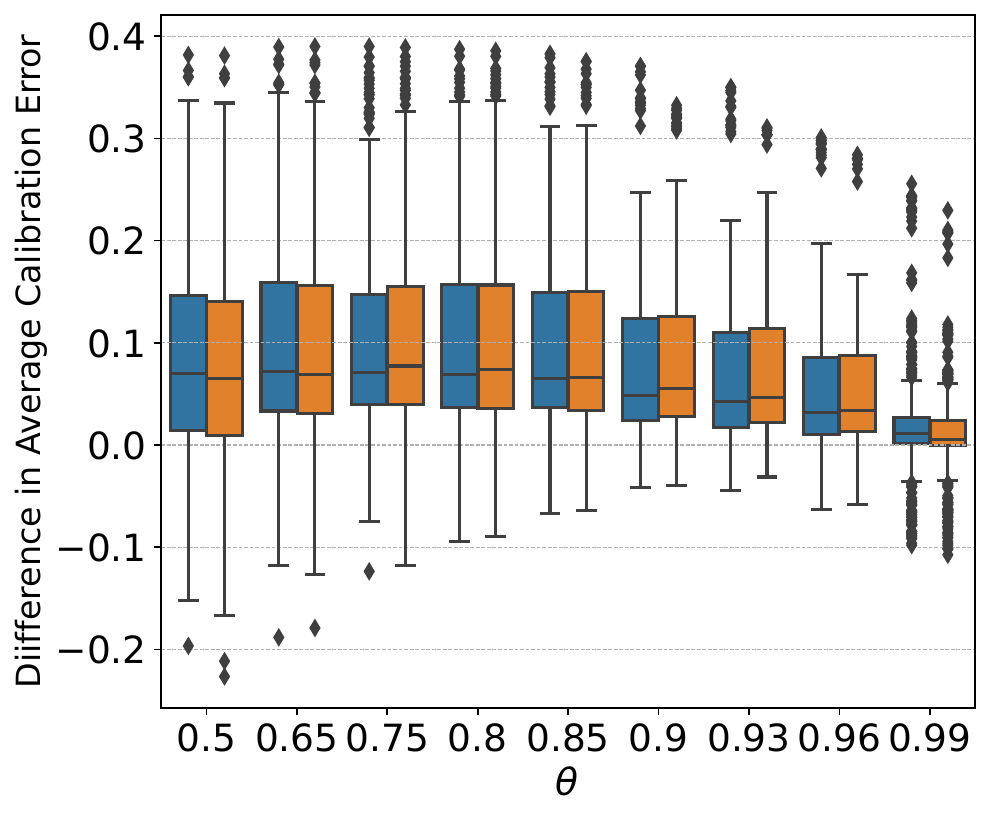}
    \end{subfigure}
    \hfill
    \begin{subfigure}[b]{0.49\linewidth}
        \centering
        \includegraphics[width=\linewidth]{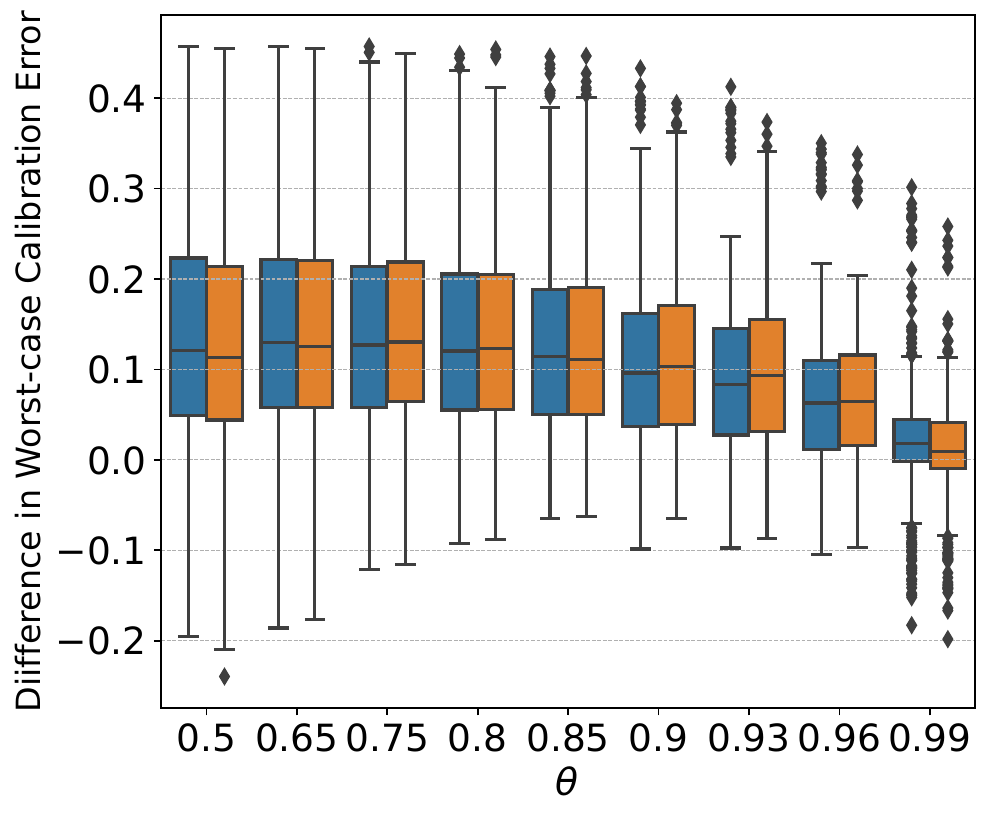}
    \end{subfigure}
    
    \vspace{0.2cm} 

    \begin{subfigure}[b]{0.49\linewidth}
        \centering
        \includegraphics[width=\linewidth]{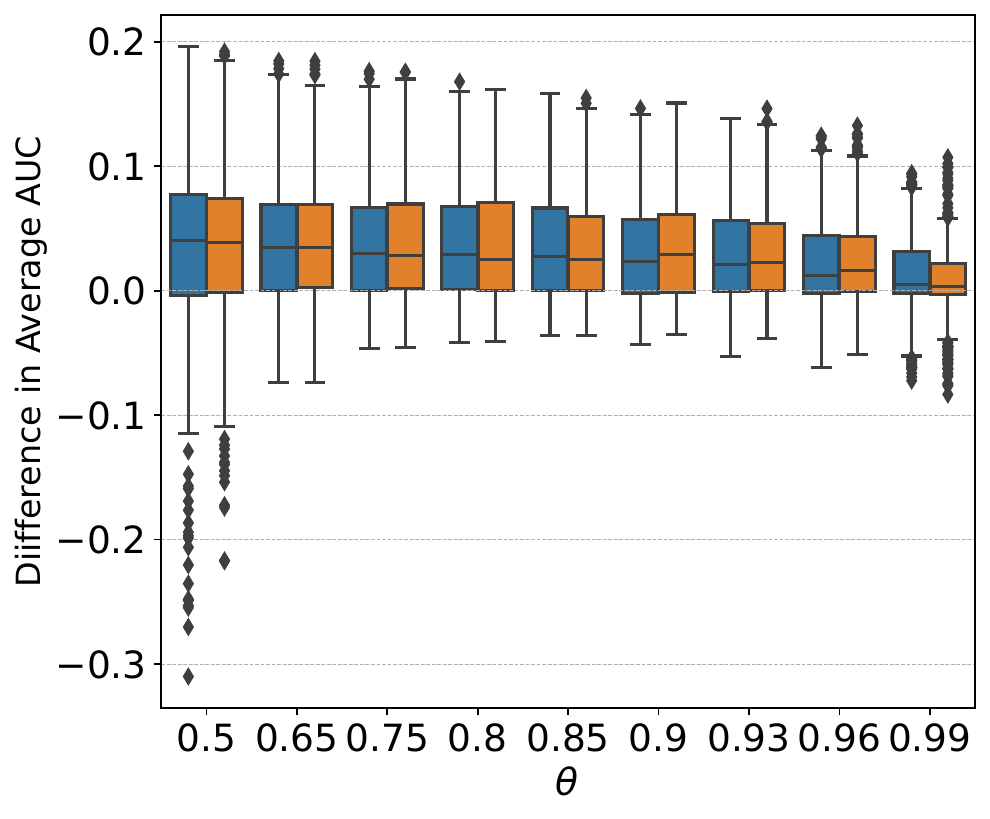}
    \end{subfigure}
    \hfill
    \begin{subfigure}[b]{0.49\linewidth}
        \centering
        \includegraphics[width=\linewidth]{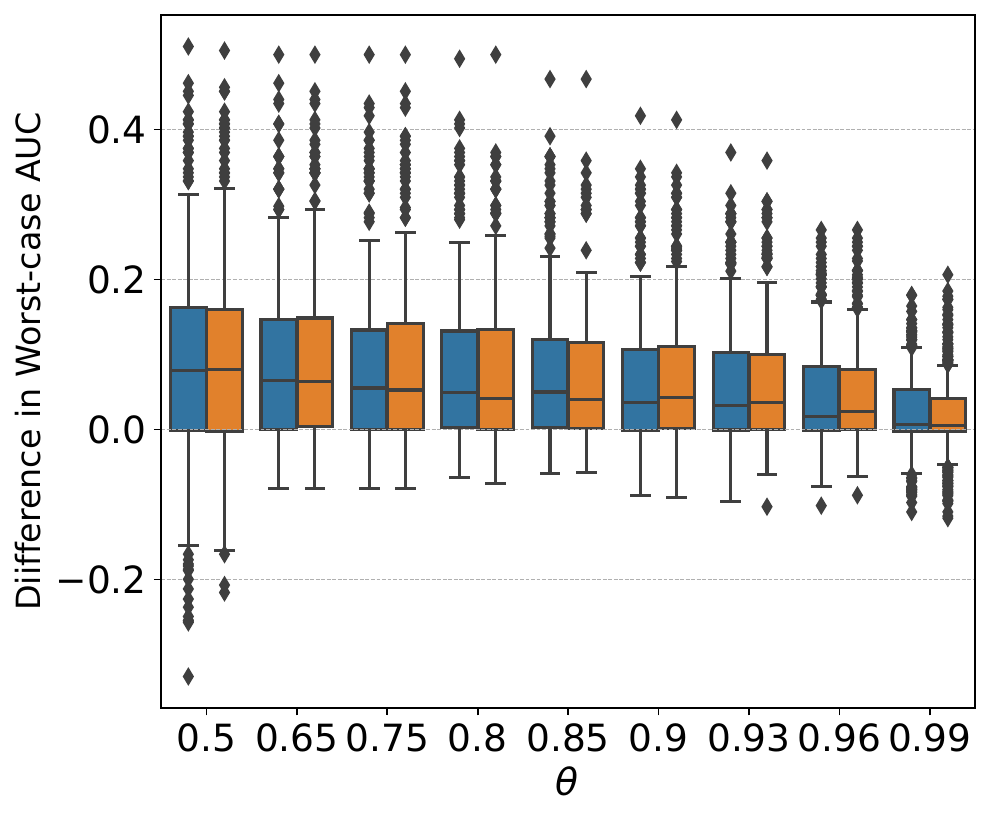}
    \end{subfigure}

    \caption{Overview of performance improvement compared to distributionally robust logistic regression with all weight parameters set to 1 in terms of calibration error and AUC across different levels of robustness $\theta$ under \emph{unexpected} perturbations. The blue boxes represent our proposed models with calibrated parameters rounded to integer. The orange boxes represent  our proposed models with calibrated parameters rounded to one decimal place.}
    \label{fig:comparison with m1 un}
\end{figure*}

The distributions of performance improvements for our proposed model under unexpected perturbations, compared to lasso logistic regression and the distributionally robust logistic regression model proposed by \citet{closest}, are summarized in Figures~\ref{fig:comparison with lasso un} and \ref{fig:comparison with m1 un}, respectively. These results highlight our model's enhanced robustness and improved performance when the basic domain knowledge is not precisely estimated and our model is misspecified.

\section{PROOFS}

\subsection{Proof of Theorem \ref{convex formulation}}
\label{sec:proof of convex formulation}

The proof of Theorem \ref{convex formulation} requires the following lemma. 
\begin{lem} \label{lemma: sup}
    Consider the convex function $g_{\bm{\beta}}(\bm{x}):=\log(1+\exp(- \bm{\beta}_{\mathrm{x}}^\top \bm{x} - \alpha))$ where $\bm{\beta}_{\mathrm{x}}, \bm{x} \in \mathbb{R}^n$ and $\alpha \in \mathbb{R}$. Then, for every $\lambda >0$, the following 
		\begin{align*}
		\sup\limits _{\bm{x} \in \mathbb{R}^n} & g_{\bm{\beta}}(\bm{x}) - \lambda \| \Gamma_x(\bm{x}-\hat{\bm{x}})\| =\begin{cases}
		g_{\bm{\beta}}(\hat{\bm{x}}) \quad &	\text{\emph{if }} \|\Gamma^{-1}_{\mathrm{x}}\bm{\beta}_{\mathrm{x}}\|_* \leq \lambda, \\
		+\infty & \text{\emph{otherwise}}
		\end{cases}
		\end{align*}
holds, where \( \| \cdot \|_* \) denotes the dual norm of \( \| \cdot \| \), specifically
\[
\|\Gamma_{\mathrm{x}}^{-1} \bm{\beta}_{\mathrm{x}}\|_* := \sup_{\| \bm{x} \| \leq 1} (\Gamma_{\mathrm{x}}^{-1} \bm{\beta}_{\mathrm{x}})^\top \bm{x},
\]
and \( \Gamma_{\mathrm{x}} \in \mathbb{R}^{n \times n} \) is a diagonal matrix with positive diagonal elements \( \gamma_j > 0 \) for \( j \in [n] \).
\end{lem}
Lemma \ref{lemma: sup} generalizes Lemma 1 of \citet{NIPS2015} by accounting for the weight parameter $\gamma_{j}$ for each numerical feature $j \in [n]$ and the constant $\alpha$ in the log-loss function. 

\begin{customproof}[Proof of Lemma \ref{lemma: sup}] 
To process the constant $\alpha$, we define $\bm{x}' = \bm{x} + \bm{c}$ where $\bm{c} \in \mathbb{R}^n$ such that $\bm{c}^\top\bm{\beta}_{\mathrm{x}} = \alpha$ and focus on showing that the following 
\begin{equation} \label{eq:remove sup}
\begin{aligned}
		\sup\limits _{\bm{x}' \in \mathbb{R}^n} & h_{\bm{\beta}}(\bm{x}') - \lambda \| \Gamma_x(\bm{x}'-\hat{\bm{x}}')\| =\begin{cases}
		h_{\bm{\beta}}(\hat{\bm{x}}') \quad &	\text{if } \|\Gamma^{-1}_{\mathrm{x}}\bm{\beta}_{\mathrm{x}}\|_* \leq \lambda, \\
		+\infty & \text{otherwise}
		\end{cases}
\end{aligned}
\end{equation}
holds where $h_{\bm{\beta}}(\bm{x}'):=\log(1+\exp(- \bm{\beta}_{\mathrm{x}}^\top \bm{x}'))$

Following the proof of Lemma 1 of \citet{NIPS2015}, we express $h_{\bm{\beta}}(\bm{x}') - \lambda \| \Gamma_x(\bm{x}' - \hat{\bm{x}}')\|$ as an upper envelop of infinityly many affine functions. 
We define $f(t):=\log(1+\exp(- t))$ and its conjugate function of $f(t)$ is
\begin{align*}
f^*(\tau)= 
\left\{ \begin{array}{l@{~}l}
\tau\log(\tau)+(1-\tau)\log(1-\tau) \quad & \text{if } \tau \in [0,1], \\[1ex]
+\infty & \text{otherwise.}
\end{array}\right.
\end{align*}
Based on strong Lagrangian duality, the conjugate of $h_{\bm{\beta}}(\bm{x}')=f(\bm{\beta}_{\mathrm{x}}^\top \bm{x}')$ is 
\begin{align*}
h_{\bm{\beta}}^*(\bm{\zeta})=
\begin{cases}
\inf\limits_{0\leq\tau\leq1} f^*(\tau) \quad & \text{if } \bm{\zeta} =\tau \bm{\beta}_{\mathrm{x}}, \\
+\infty & \text{otherwise.}
\end{cases}
\end{align*}

As the logloss function $h_{\bm{\beta}}(\bm{x}')$ is convex and continuous, it equals its bi-conjugate: 
\begin{align*}
h_{\bm{\beta}}(\bm{x}') = h^{**}_{\bm{\beta}}(\bm{x}') & = \sup_{\bm{\zeta} \in \mathbb{R}^n} \langle \bm{\zeta}, \bm{x}' \rangle - h_{\bm{\beta}}^{*}(\bm{\zeta}) \notag \\
& = \sup\limits_{0 \leq \tau \leq 1} 
\langle \tau\bm{\beta}_{\mathrm{x}}, \bm{x}' \rangle -f^*(\tau).
\end{align*}
Using this representation, we obtain
\begin{align*}
\quad \sup\limits_{\bm{x}'\in\mathbb{R}^n} h_{\bm{\beta}}(\bm{x}') - \lambda \|\Gamma_{x}(\bm{x}' - \hat{\bm{x}}')\| & = \sup\limits_{\bm{x}'\in\mathbb{R}^n} h^{**}_{\bm{\beta}}(\bm{x}') - \lambda \|\Gamma_{x}(\bm{x}' - \hat{\bm{x}}')\| \notag \\
& = \sup\limits_{0 \leq \tau \leq 1} \sup_{\bm{x}' \in\mathbb{R}^n}
\langle \tau\bm{\beta}_{\mathrm{x}}, \bm{x}' \rangle -f^*(\tau) - \lambda \|\Gamma_{x}(\bm{x}'- \hat{\bm{x}}')\| \notag \\
&=\sup\limits_{0 \leq \tau \leq 1} \sup_{\bm{x}' \in \mathbb{R}^n} 
\langle \tau\bm{\beta}_{\mathrm{x}}, \bm{x}' \rangle -f^*(\tau) - \sup\limits_{\|q\|_*\leq\lambda} \langle q , \Gamma_{x}(\bm{x}'- \hat{\bm{x}}') \rangle \notag \\
&=\sup\limits_{0 \leq \tau \leq 1} \sup_{\bm{x}' \in \mathbb{R}^n} 
\langle \tau\bm{\beta}_{\mathrm{x}}, \bm{x}' \rangle -f^*(\tau) - \sup\limits_{\|q\|_*\leq\lambda} \langle \Gamma_{x} q , \bm{x}'- \hat{\bm{x}}' \rangle \notag \\
&=\sup\limits_{0 \leq \tau \leq 1} \sup_{\bm{x}' \in \mathbb{R}^n} \inf\limits
_{\|q\|_* \leq \lambda}
\langle \tau\bm{\beta}_{\mathrm{x}}, \bm{x}' \rangle -f^*(\tau) - \langle \Gamma_{x} q , \bm{x}'- \hat{\bm{x}}' \rangle \notag \\
&=\sup\limits_{0 \leq \tau \leq 1} \inf\limits_{\|q\|_* \leq \lambda}
\sup_{\bm{x}' \in \mathbb{R}^n} \langle \tau\bm{\beta}_{\mathrm{x}} - \Gamma_{x} q , \bm{x}' \rangle - f^*(\tau) +  \langle \Gamma_{x} q , \hat{\bm{x}}' \rangle, \notag
\end{align*}
where the third equality follows from the definition of the dual norm. Explicitly evaluating the maximization over $\bm{x}'$ shows that the above expression is equivalent to 	
\begin{align*}
&\begin{cases}
\sup\limits_{0 \leq \tau \leq 1} \inf\limits_{\|q\|_* \leq \lambda} &
-f^*(\tau) + \langle \Gamma_{x} q , \hat{\bm{x}}' \rangle \\
\text{~ s.t.} &    \tau \bm{\beta}_{\mathrm{x}} - \Gamma_{x} q= 0
\end{cases} \notag \\
=&\begin{cases}
\sup\limits_{0 \leq \tau \leq 1} 	-f^*(\tau) + \langle \tau \bm{\beta}_{\mathrm{x}}, \hat{\bm{x}}' \rangle & \text{if }\sup\limits_{0 \leq \tau \leq 1} \| \tau \Gamma_{x}^{-1} \bm{\beta}_{\mathrm{x}} \|_* \leq \lambda, \\
+\infty & \text{otherwise.}
\end{cases} \notag
\end{align*}
We conclude that equation \eqref{eq:remove sup} holds. By expressing $\bm{x}'$ in terms of $\bm{x} + \bm{c}$, we have 
		\begin{align*}
		\sup\limits _{\bm{x} \in \mathbb{R}^n} & g_{\bm{\beta}}(\bm{x}) - \lambda \| \Gamma_x(\bm{x}-\hat{\bm{x}})\| =\begin{cases}
		g_{\bm{\beta}}(\hat{\bm{x}}) \quad &	\text{if } \|\Gamma^{-1}_{\mathrm{x}}\bm{\beta}_{\mathrm{x}}\|_* \leq \lambda, \\
		+\infty & \text{otherwise.}
		\end{cases}
		\end{align*}
Therefore, Lemma \ref{lemma: sup} holds. 
\end{customproof}

\begin{customproof}[Proof of Theorem \ref{convex formulation}]
    The theorem can be proven along the lines of the proof of Theorem 1  by \citet{closest} if we leverage Lemma \ref{lemma: sup}. Problem \eqref{eq:original formulation} can be reformulated as 
    \begin{equation*}
\begin{array}{l@{\,}l@{\,}l}
        \displaystyle \mathop{\text{minimize}}_{\lambda, \bm{r}} & \displaystyle \lambda \epsilon + \dfrac{1}{N}\sum_{i=1}^N r_i \\[6mm]
        \displaystyle \text{subject to} & \displaystyle \underset{\bm{x} \in \mathbb{R}^n}{\sup} \left\{ l_{\bm{\beta}}(\bm{x}, \bm{z}, +1) - \lambda \sum_{j \in [n]} \gamma_{j} |x_j - x^i_j| \right\} - \lambda \kappa  \indicate{y^i \neq 1} - \lambda \sum_{\ell \in [m]} \delta_{\ell} \indicate{\bm{z}_{\ell} \neq \bm{z}^i_{\ell}} \leq r_i, \,  \forall i \in [N], \bm{z} \in \mathcal{C} \\ [7mm]
        & \displaystyle \underset{\bm{x} \in \mathbb{R}^n}{\sup} \left\{ l_{\bm{\beta}}(\bm{x}, \bm{z}, -1) - \lambda \sum_{j \in [n]} \gamma_{j} |x_j - x^i_j| \right\} - \lambda \kappa  \indicate{y^i \neq -1} - \lambda \sum_{\ell \in [m]} \delta_{\ell} \indicate{\bm{z}_{\ell} \neq \bm{z}^i_{\ell}}  \leq r_i, 
        \, \forall i \in [N], \bm{z} \in \mathcal{C} \\ [6mm]
        & \displaystyle \lambda \geq 0, \;\; \bm{r} \in \mathbb{R}^N
    \end{array}
\end{equation*}
The dual norm of $l_1$ norm is $l_{\infty}$ norm. 
Applying Lemma \ref{lemma: sup} to the suprema of the above problem and choosing $l_1$ norm, we get
    \begin{equation*}
    \mspace{-35mu}
    \begin{array}{l@{\quad}l@{\qquad}l}
        \displaystyle \mathop{\text{minimize}}_{\lambda, \bm{r}} & \displaystyle \lambda \epsilon + \dfrac{1}{N}\sum_{i=1}^N r_i \\[4mm]
        \displaystyle \text{subject to} & \displaystyle  l_{\bm{\beta}}(\bm{x}^i, \bm{z}, +1) - \lambda \kappa \cdot \indicate{y^i \neq 1} - \lambda \sum_{\ell \in [m]} \delta_{\ell} \indicate{\bm{z}_{\ell} \neq \bm{z}^i_{\ell}} \leq r_i, \quad \forall i \in [N],\; \bm{z} \in \mathcal{C} \\ [5mm]
        & \displaystyle  l_{\bm{\beta}}(\bm{x}^i, \bm{z}, -1) - \lambda \kappa \cdot \indicate{y^i \neq -1} - \lambda \sum_{\ell \in [m]} \delta_{\ell} \indicate{\bm{z}_{\ell} \neq \bm{z}^i_{\ell}}  \leq r_i, \quad \forall i \in [N],\; \bm{z} \in \mathcal{C} \\ [5mm]
        & \max_{j \in [n]} | \gamma_{j}^{-1} \beta_{\mathrm{x}j}| \leq \lambda \\[3mm]
        & \displaystyle \lambda \geq 0, \;\; \bm{r} \in \mathbb{R}^N.
    \end{array}
\end{equation*}
Based on the property of maximization problem, this formulation is equivalent to 
    \begin{equation*}
    \mspace{-35mu}
    \begin{array}{l@{\quad}l@{\qquad}l}
        \displaystyle \mathop{\text{minimize}}_{\lambda, \bm{r}} & \displaystyle \lambda \epsilon + \dfrac{1}{N}\sum_{i=1}^N r_i \\[4mm]
        \displaystyle \text{subject to} & \displaystyle  l_{\bm{\beta}}(\bm{x}^i, \bm{z}, +1) - \lambda \kappa \cdot \indicate{y^i \neq 1} - \lambda \sum_{\ell \in [m]} \delta_{\ell} \indicate{\bm{z}_{\ell} \neq \bm{z}^i_{\ell}} \leq r_i, \quad \forall i \in [N],\; \bm{z} \in \mathcal{C} \\ [5mm]
        & \displaystyle  l_{\bm{\beta}}(\bm{x}^i, \bm{z}, -1) - \lambda \kappa \cdot \indicate{y^i \neq -1} - \lambda \sum_{\ell \in [m]} \delta_{\ell} \indicate{\bm{z}_{\ell} \neq \bm{z}^i_{\ell}}  \leq r_i, 
        \quad \forall i \in [N],\; \bm{z} \in \mathcal{C} \\ [5mm]
        & |\gamma_{j}^{-1} \beta_{\mathrm{x}j}| \leq \lambda, 
        \quad \forall j \in [n] \\[3mm]
        & \displaystyle \lambda \geq 0, \;\; \bm{r} \in \mathbb{R}^N.
    \end{array}
\end{equation*}

For $y^i = +1$, the identities
\begin{align*}
    l_{\bm{\beta}}(\bm{x}^i, \bm{z}, +1) - \lambda \kappa \cdot \indicate{y^i \neq 1} \;\; &= \;\; l_{\bm{\beta}}(\bm{x}^i, \bm{z}, y^i) \\
    l_{\bm{\beta}}(\bm{x}^i, \bm{z}, -1) - \lambda \kappa \cdot \indicate{y^i \neq -1} \;\; &= \;\; l_{\bm{\beta}}(\bm{x}^i, \bm{z}, -y^i) - \lambda \kappa
\end{align*}
hold; for $y^i = -1$, the identities
\begin{align*}
    l_{\bm{\beta}}(\bm{x}^i, \bm{z}, +1) - \lambda \kappa \cdot \indicate{y^i \neq 1} \;\; &= \;\; l_{\bm{\beta}}(\bm{x}^i, \bm{z}, -y^i) - \lambda \kappa\\
    l_{\bm{\beta}}(\bm{x}^i, \bm{z}, -1) - \lambda \kappa \cdot \indicate{y^i \neq -1} \;\; &= \;\; l_{\bm{\beta}}(\bm{x}^i, \bm{z}, y^i)
\end{align*}
hold as well. we can simplify above formulation to
    \begin{equation*}
    \mspace{-35mu}
    \begin{array}{l@{\quad}l@{\qquad}l}
        \displaystyle \mathop{\text{minimize}}_{\lambda, \bm{r}} & \displaystyle \lambda \epsilon + \dfrac{1}{N}\sum_{i=1}^N r_i \\[4mm]
        \displaystyle \text{subject to} & \displaystyle  l_{\bm{\beta}}(\bm{x}^i, \bm{z}, y^i)  - \lambda \sum_{\ell \in [m]} \delta_{\ell} \indicate{\bm{z}_{\ell} \neq \bm{z}^i_{\ell}} \leq r_i, \quad \forall i \in [N],\; \bm{z} \in \mathcal{C} \\ [5mm]
        & \displaystyle  l_{\bm{\beta}}(\bm{x}^i, \bm{z}, -y^i) - \lambda \kappa - \lambda \sum_{\ell \in [m]} \delta_{\ell} \indicate{\bm{z}_{\ell} \neq \bm{z}^i_{\ell}}  \leq r_i, 
        \quad \forall i \in [N],\; \bm{z} \in \mathcal{C} \\ [5mm]
        & |\gamma_{j}^{-1} \beta_{\mathrm{x}j}| \leq \lambda, \quad \forall j \in [n]\\[3mm]
        & \displaystyle \lambda \geq 0, \;\; \bm{r} \in \mathbb{R}^N
    \end{array}
\end{equation*}
Since we do not account for perturbations in the labels, we set 
$\kappa$ to infinity, causing the second set of constraints to hold trivially. Removing these constraints and incorporating this problem into the overall optimization problem, we can get  problem \eqref{eq:convex reformulation}
    \begin{equation*}
    \mspace{-35mu}
    \begin{array}{l@{\quad}l@{\qquad}l}
        \displaystyle \mathop{\text{minimize}}_{\lambda, \bm{r}, \bm{\beta}} & \displaystyle \lambda \epsilon + \dfrac{1}{N}\sum_{i=1}^N r_i \\[4mm]
        \displaystyle \text{subject to} & \displaystyle  l_{\bm{\beta}}(\bm{x}^i, \bm{z}, y^i)  - \lambda \sum_{\ell \in [m]} \delta_{\ell} \indicate{\bm{z}_{\ell} \neq \bm{z}^i_{\ell}} \leq r_i, \quad \forall i \in [N],\; \bm{z} \in \mathcal{C} \\ [5mm]
        & |\gamma_{j}^{-1} \beta_{\mathrm{x}j}| \leq \lambda, \quad \forall j \in [n] \\[3mm]
        & \displaystyle \lambda \geq 0, \;\; \bm{r} \in \mathbb{R}^N, \; \; {\bm \beta} = (\beta_0, \bm{\beta}_{\mathrm{x}}, \bm{\beta}_{\mathrm{z}}) \in \mathbb{R}^{1+n+c}
    \end{array}
\end{equation*}
\end{customproof}

\textbf{Equivalent formulation of problem \eqref{eq:convex reformulation} as an exponential cone problem.} We can convert a logarithm constraint $\log(1 + \exp(c)) \leq a$ into $\exp(-a) + \exp(c - a) \leq 1$ through exponentiation operations.
By introducing the auxiliary variables $u, v$, we can reformulate this constraint as
\begin{equation}\label{eq:logarithm constraints}
\begin{aligned}
    & u + v \leq 1 \\
    & \exp(-a) \leq u \\
    & \exp(c - a) \leq v
\end{aligned}
\end{equation}
The exponential cone is defined as
\begin{align*}
    \mathcal{K}_{\exp} := \mathrm{cl} \left(\left\{(a,b,c) \ : \ a \geq b \exp(c / b), \ a > 0, \ b > 0 \right\}\right) \subset \mathbb{R}^3,
\end{align*}
where $\mathrm{cl}(\cdot)$ denotes the closure. Therefore, we can rewrite constraints \eqref{eq:logarithm constraints} as
\begin{equation*}
\begin{aligned}
        &u + v \leq 1 \\
        &(u, 1, -a) \in \mathcal{K}_{\exp} \\
        &(v, 1, c-a) \in \mathcal{K}_{\exp}
\end{aligned}
\end{equation*}
For each $(i, \bm{z}) \in [N] \times \mathcal{C}$, the constraint in problem \eqref{eq:convex reformulation}
\[
\log \left( 1 + \exp \left( -y^i \left( \bm{\beta}_{\mathrm{x}}^\top \bm{x}^i + \bm{\beta}_{\mathrm{z}}^\top \bm{z} \right) \right) \right)  - \lambda \sum_{\ell \in [m]} \delta_{\ell} \indicate{\bm{z}_{\ell} \neq \bm{z}^i_{\ell}} \leq r_i
\]
can be converted into 
\begin{equation*}
\begin{aligned}
        &u + v \leq 1 \\
        &(u, 1, - \lambda \sum_{\ell \in [m]} \delta_{\ell} \indicate{\bm{z}_{\ell} \neq \bm{z}^i_{\ell}} - r_i) \in \mathcal{K}_{\exp} \\
        &(v, 1, -y^i \left( \bm{\beta}_{\mathrm{x}}^\top \bm{x}^i + \bm{\beta}_{\mathrm{z}}^\top \bm{z} \right) - \lambda \sum_{\ell \in [m]} \delta_{\ell} \indicate{\bm{z}_{\ell} \neq \bm{z}^i_{\ell}} - r_i) \in \mathcal{K}_{\exp}
\end{aligned}
\end{equation*}
We can formulate problem \eqref{eq:convex reformulation} in the form of exponential cone problem
    \begin{equation*}
        \mspace{-20mu}
        \begin{array}{l@{\quad}l}
            \displaystyle \mathop{\text{minimize}}_{\lambda, \boldsymbol{r}, \bm{\beta}, \bm{u}, \bm{v}} & \displaystyle \lambda \epsilon + \dfrac{1}{N}\sum_{i \in [N]} r_i \\[6mm]
            \displaystyle \raisebox{9.5mm}{$\mspace{1mu}$ \text{subject to}} &
            \mspace{-10mu} \left. \begin{array}{l}
                \displaystyle u_{i\bm{z}} + v_{i\bm{z}} \leq 1 \\ [2mm]
                \displaystyle (u_{i\bm{z}}, 1, - \lambda \sum_{\ell \in [m]} \delta_{\ell} \indicate{\bm{z}_{\ell} \neq \bm{z}^i_{\ell}} - r_i) \in \mathcal{K}_{\exp} \\
                \displaystyle (v_{i\bm{z}}, 1, -y^i \left( \bm{\beta}_{\mathrm{x}}^\top \bm{x}^i + \bm{\beta}_{\mathrm{z}}^\top \bm{z} \right) - \lambda \sum_{\ell \in [m]} \delta_{\ell} \indicate{\bm{z}_{\ell} \neq \bm{z}^i_{\ell}} - r_i)) \in \mathcal{K}_{\exp}
            \end{array}
            \mspace{2mu} \right\} \displaystyle \forall i \in [N], \; \bm{z} \in \mathcal{C} \\[6mm]
            & |\gamma_{j}^{-1} \beta_{\mathrm{x}j}| \leq \lambda, \;\quad \forall j \in [n] \\[3mm]
            & \lambda \geq 0, \, \bm{r} \in \mathbb{R}^N, 
            \, {\bm \beta} = (\beta_0, \bm{\beta}_{\mathrm{x}}, \bm{\beta}_{\mathrm{z}}) \in \mathbb{R}^{1+n+c} \\ [2mm]
            & \displaystyle (u_{i\bm{z}}, v_{i\bm{z}}) \in \mathbb{R}^2, \; (i,\bm{z}) \in [N] \times \mathcal{C}
        \end{array}
    \end{equation*}

\subsection{Proof of Correctness of Algorithm \ref{alg:dp}} \label{sec: proof of dp algorithm}
\begin{proof}
    We prove the correctness of Algorithm \ref{alg:dp} using mathematical induction, showing that we can calculate $g^i(k,d)$ correctly for each $(k,d) \in \mathcal{S}^i$  by equations \eqref{eq:bellman g k=1m}-\eqref{eq:bellman g k=m+1}.

\textbf{Base Case:}

For \( k = 0 \), we have $g^i(0,0) = 0$.
This is given directly by the definition, and since there are no categorical features to consider, the base case holds trivially.

\textbf{Inductive Step:}

\textbf{For each \( \boldsymbol{k \in [m]} \),}
assume that the value \( g^i(k-1, d') \) has been correctly calculated for each \((k-1, d') \in \mathcal{S}^i \).  Then, for each state $(k,d) \in \mathcal{S}^i$, the function \( g^i(k, d) \) is defined as \eqref{eq:subproblem k=1m}, which is:
\begin{equation*}
    \begin{array}{ccl}
        g^i(k,d) \; := & \underset{\{\bm{z}_{\ell}\}_{\ell \in [k]}}{\text{max}} &  \displaystyle  -y^i \sum_{\ell \in [k]} \bm{\beta}_{\mathrm{z}\ell}^\top \bm{z}_{\ell}   \\[5mm]
        & \text{s.t.}  & {\bm{z}_{\ell} \in \mathcal{C}_{\ell}, \; \ell \in [k]} \\[2mm]
        && \displaystyle \sum_{\ell \in [k]} \delta_{\ell} \indicate{\bm{z}_{\ell} \neq \bm{z}^i_{\ell}} = d
    \end{array}
\end{equation*}
We can decompose this optimization problem by dividing decision variables $\{\bm{z}_{\ell}\}_{\ell \in [k]}$ into \( \bm{z}_k \) and \( \{\bm{z}_{\ell}\}_{\ell \in [k-1]}\):
\begin{equation*}
    \begin{array}{ccl}
        g^i(k,d) \; = & \underset{\bm{z}_k \in \mathcal{C}_k}{\text{max}} & -y^i \bm{\beta}_{\mathrm{z}k}^\top \bm{z}_k + \underset{\{\bm{z}_{\ell}\}_{\ell \in [k-1]}}{\text{max}} \left[ -y^i \sum_{\ell \in [k-1]} \bm{\beta}_{\mathrm{z}\ell}^\top \bm{z}_{\ell} \right]    \\[5mm]
        & \text{s.t.}   & (k-1, d - \delta_k \indicate{\bm{z}_k \neq \bm{z}^i_k}) \in \mathcal{S}^i \\[2mm]
        && {\bm{z}_{\ell} \in \mathcal{C}_{\ell}, \; \ell \in [k-1]} \\[2mm]
        && \displaystyle \sum_{\ell \in [k-1]} \delta_{\ell} \indicate{\bm{z}_{\ell} \neq \bm{z}^i_{\ell}} = d - \delta_k \indicate{\bm{z}_k \neq \bm{z}^i_k} 
    \end{array}
\end{equation*}
The first constraint ensures that $\bm{z}_k$ is feasible for subproblem \eqref{eq:subproblem k=1m} at state $(k,d)$. The second and the third constraints define the feasible region for subproblem \eqref{eq:subproblem k=1m} at state $(k-1,d - \delta_k \indicate{\bm{z}_k \neq \bm{z}^i_k})$ when choosing $\bm{z}_k \in \mathcal{C}_k$. 
By the inductive hypothesis, the inner maximization over \( \{\bm{z}_{\ell}\}_{\ell \in [k-1]}\) yields \( g^i(k-1, d - \delta_k \indicate{\bm{z}_k \neq \bm{z}^i_k})\). Therefore:
\begin{equation*}
\label{eq:bellman_recursion}
g^i(k, d) = \underset{\bm{z}_k \in \mathcal{F}_{kd}}{\text{max}} \left\{ -y^i \bm{\beta}_{\mathrm{z}k}^\top \bm{z}_k + g^i(k-1, d - \delta_k \indicate{\bm{z}_k \neq \bm{z}^i_k}) \right\}
\end{equation*}
where
\[
\mathcal{F}_{kd} = \left\{ \bm{z}_k \in \mathcal{C}_k \; \middle| \; (k-1, d - \delta_k \indicate{\bm{z}_k \neq \bm{z}^i_k}) \in \mathcal{S}^i \right\}.
\]
This is exactly the Bellman equation \eqref{eq:bellman g k=1m}.

\textbf{For \( \boldsymbol{k = m+1} \),} we must account for the numerical feature component, the logistic function, and current solution to the relaxed problem of \eqref{eq:convex reformulation} that are not captured in previous subproblems. This ensures that \( g^i(m+1, 0) \) represents the optimal objective value of problem \eqref{eq:identification}. We can compare the constraint violations of the optimal \( \bm{z} \) for subproblem \eqref{eq:subproblem k=1m} across all $(m,d) \in S^i_1$ and select the one with the largest violation as the most violated constraint. 
Using the values of \( g^i(m, d) \), we have:
\begin{equation*}
\label{eq:final_step}
\begin{aligned}
g^i(m+1, 0) = \underset{d}{\text{max}} \; & \log\left(1 + \exp\left(-y^i \bm{\beta}_{\mathrm{x}}^\top \bm{x}^i + g^i(m, d)\right)\right) - \lambda d - r_i \\
\text{s.t.} \; & (m, d) \in \mathcal{S}^i_1
\end{aligned}
\end{equation*}
This aligns with equation \eqref{eq:bellman g k=m+1}, allowing us to compute \( g^i(m+1, 0) \) using the previously calculated \( g^i(m, d) \).
Therefore, we can calculate $g^i(k,d)$ for each $(k,d) \in \mathcal{S}^i$ by equations \eqref{eq:bellman g k=1m}-\eqref{eq:bellman g k=m+1}.
\end{proof}

\subsection{Proof of Lemma \ref{dp to path}}
\label{sec:Proof of Lemma dp to path}
\begin{proof}
For clarity in the following proof, we explicitly restate the dynamic programming subproblems used to solve problem \eqref{eq:constraint reformulation}, along with the function $h^i$, which represents their optimal objective values, as defined in Section \ref{sec:graph}. This restatement is provided without introducing new definitions.

If $k=0$, for state $(0,0)$,
\begin{equation} \label{eq:subproblem1}
    h^i(0,0):= 0.
\end{equation}

If $k \in [m]$, for each state $(k,d) \in \mathcal{S}^i_1$,
\begin{equation}\label{eq:subproblem2}
    \begin{array}{ccl}
        h^i(k,d) := &\underset{\{\bm{z}_{\ell}\}_{\ell \in [k]}}{\text{max}} &  \displaystyle  -y^i \sum_{\ell \in [k]} \bm{\beta}_{\mathrm{z}\ell}^\top \bm{z}_{\ell}   \\[5mm]
        & \text{s.t.}  & {\bm{z}_{\ell} \in \mathcal{C}_{\ell}, \; \ell \in [k]} \\[2mm]
        && \displaystyle \sum_{\ell \in [k]} \delta_{\ell} \indicate{\bm{z}_{\ell} \neq \bm{z}^i_{\ell}} = d.
    \end{array}
\end{equation}
If $k = m+1$, for state $(m+1,0)$,
\begin{equation}\label{eq:subproblem3}
\begin{aligned}
    h^i(m+1,0) := \max_{\bm{z} \in \mathcal{C}}  & \; -y^i \bm{\beta}_{\mathrm{z}}^\top \bm{z} - \log \Big( -1 +\exp \Big( r_i + \lambda \sum_{\ell \in [m]} \delta_{\ell} \indicate{\bm{z}_{\ell} \neq \bm{z}^i_{\ell}} \Big) \Big).
\end{aligned}
\end{equation}
These subproblems can be solved by Bellman equations \eqref{eq:bellman h k=1m}-\eqref{eq:bellman h k=m+1}. The proof follows the proof of Algorithm \ref{alg:dp} and is omitted for the sake of brevity. 

Next, having established the equivalence between problem \eqref{eq:constraint reformulation} and the dynamic programming approach, we proceed to link the dynamic programming method with the longest path problem in the corresponding graph. Due to the correspondence between the vertex set \(\mathcal{V}^i\) and the state space \(\mathcal{S}^i\) for each data point \(i\), we assert that in each graph \(\mathcal{G}^i\), the longest path from the source \((0,0)\) to any vertex \((k,d) \in \mathcal{V}^i\) corresponds to the optimal solution of subproblems \eqref{eq:subproblem1}-\eqref{eq:subproblem3}, and the sum of the weights along this path equals \(h^i(k,d)\). We prove this via mathematical induction.

Based on the graph structure, all vertices with the same \( k = 0, 1, \dots, m+1 \) are grouped into the same layer \( k \). For each vertex \( (k,d) \in \mathcal{V}^i \), where \( k = 1, \dots, m+1 \), there is at least one arc targeting \( (k,d) \) from a vertex in the previous layer \( k-1 \). Thus, for any vertex \( (k,d) \in \mathcal{V}^i \setminus \{(0,0)\} \), there always exists a path from source \( (0,0) \) to \( (k,d) \).

For \( k \in [m] \), since every arc is from a vertex in layer \( k-1 \) to a vertex in layer \( k \) and corresponds to a categorical feature realization \( \bm{z}_k \in \mathcal{C}_{k} \) , any path from the source \( (0,0) \) to a vertex \( (k,d) \in \mathcal{V}^i \) can be described as a sequence of realizations of the first \( k \) categorical features. The path from the source \( (0,0) \) to itself is represented by the empty sequence \( () \).

We define the set of all possible paths from the source \( (0,0) \) in the graph \( \mathcal{G}^i \) as follows:
\[
\Pi^i := \left\{ (\bm{z}_{\ell})_{\ell \in [k]} \mid \bm{z}_{\ell} \in \mathcal{C}_{\ell}, \; k \in [m] \right\} \cup \left\{() \right\}.
\]

We define the longest path from source $(0,0)$ to vertex $(k,d) \in \mathcal{V}^i$ as the function $\pi^i: \mathcal{V}^i \rightarrow \Pi^i$.

\textbf{Base Case:}

For \(k = 0\), we have \(h^i(0,0) = 0\) and $\pi^i(0,0) = ()$ as given. 

\textbf{Inductive Step:}

\textbf{For each \( \boldsymbol{k \in [m]} \),} assume that for all vertices \((k-1, d') \in \mathcal{V}^i\), the longest path from \((0,0)\) to \((k-1, d')\), denoted $\pi^i(k-1,d')$, corresponds to the optimal solution of subproblem \eqref{eq:subproblem1} or \eqref{eq:subproblem2} at state \((k-1, d')\), and the sum of the weights along this path equals \(h^i(k-1, d')\).

From equation \eqref{eq:bellman h k=1m}, we have:
\[
h^i(k, d) = \max_{\bm{z}_k \in \mathcal{F}_{kd}} \left\{ -y^i \bm{\beta}_{\mathrm{z}k}^\top \bm{z}_k + h^i\left(k-1, d - \delta_k \mathbb{I}\{\bm{z}_k \neq \bm{z}^i_k\}\right) \right\}.
\]
In the graph \(\mathcal{G}^i\), for each \(\bm{z}_k \in \mathcal{F}_{kd}\), there is an arc from the vertex \(\left(k-1, d - \delta_k \mathbb{I}\{\bm{z}_k \neq \bm{z}^i_k\}\right)\) to \((k, d)\) with a weight of \(-y^i \bm{\beta}_{\mathrm{z}k}^\top \bm{z}_k\).
By the induction hypothesis, \( \pi^i(\left(k-1, d - \delta_k \mathbb{I}\{\bm{z}_k \neq \bm{z}^i_k)\}\right)\) corresponds to the optimal solution of subproblem \eqref{eq:subproblem1} or \eqref{eq:subproblem2} at state $(k-1, d - \delta_k \mathbb{I}\{\bm{z}_k \neq \bm{z}^i_k\})$. 
Therefore, the longest path to vertex $(k,d)$ through vertex $(k-1, d - \delta_k \mathbb{I}\{\bm{z}_k \neq \bm{z}^i_k\})$ and arc corresponding to $\bm{z}_k$ is 
$(\pi^i(k-1, d - \delta_k \mathbb{I}\{\bm{z}_k \neq \bm{z}^i_k\}), \bm{z}_k)$ and its total weight is
$-  y^i \bm{\beta}_{\mathrm{z}k}^\top \bm{z}_k + h^i\left(k-1, d - \delta_k \mathbb{I}\{\bm{z}_k \neq \bm{z}^i_k\}\right)$.
Taking the maximum over all possible arcs corresponding to \(\bm{z}_k \in \mathcal{F}_{kd}\), we obtain the longest path to vertex $(k,d)$ and its total weight, as given by equation \eqref{eq:bellman h k=1m}. Therefore, our assertion holds for each $k \in [m]$.

\textbf{For \( \boldsymbol{k = m+1} \),} from equation \eqref{eq:bellman h k=m+1}, we have:
\[
h^i(m+1, 0) = \max_{(m, d') \in \mathcal{S}^i_1}  h^i(m, d') - \log\left( -1 + \exp(r_i + \lambda d') \right) .
\]
In the graph \(\mathcal{G}^i\), for each vertex \((m, d') \in \mathcal{V}^i\), there is an arc from \((m, d')\) to \((m+1, 0)\) with a weight of \(- \log\left( -1 + \exp(r_i + \lambda d') \right)\). 
By the induction hypothesis, \(\pi^i(m, d')\) corresponds to the optimal solution of subproblem \eqref{eq:subproblem2} at state $(m, d')$. Therefore, the longest path from \((0,0)\) to \((m+1, 0)\) passing through \((m, d')\) is
$\pi^i(m,d')$ and its total weight is 
$h^i(m, d') - \log\left( -1 + \exp(r_i + \lambda d') \right)$.
Taking the maximum over all possible \(d'\), we obtain the longest path to vertex $(m+1,0)$ and its total weight, as given by equation \eqref{eq:bellman h k=m+1}. Therefore, our assertion holds for $k=m+1$.

Based on our assertion, since problem \eqref{eq:constraint reformulation} corresponds to subproblem \eqref{eq:subproblem3}, Lemma \ref{dp to path} holds trivially.  
\end{proof}

\subsection{Proof of Theorem \ref{thm: graph-based reformulation}}
\label{sec: Proof of Theorem graph-based formulation}

\begin{proof}
We convert the constraint set indexed by $[N] \times \mathcal{C}$ in problem \eqref{eq:convex reformulation} into a more tractable form. 
 Given the data point indexed by $i$, 
the constraint
\begin{align*}
    r_i \geq & \; \log\left(1 + \exp\left(-y^i \left( \bm{\beta}_\mathrm{x}^\top \bm{x}^i + \bm{\beta}_\mathrm{z}^\top \bm{z} + \beta_0 \right)\right)\right) - \lambda \sum_{\ell \in [m]} \delta_{\ell} \indicate{\bm{z}_{\ell} \neq \bm{z}^i_{\ell}}
\end{align*}
must hold for all $ \bm{z}\in \mathcal{C}$. 
Each inequality can be rewritten as
\begin{align*}
   &\text{exp}\left(r_i\right)\geq \left(1+\text{exp}(-y^i(\bm{\beta}_\mathrm{x}^\top\bm{x}^i +\bm{\beta}_\mathrm{z}^\top\bm{z} + \beta_0))\right)\text{exp}\left( - \lambda \sum_{\ell \in [m]} \delta_{\ell} \indicate{\bm{z}_{\ell} \neq \bm{z}^i_{\ell}}\right)\\
   \Leftrightarrow &\text{exp}\left(r_i + \lambda \sum_{\ell \in [m]} \delta_{\ell} \indicate{\bm{z}_{\ell} \neq \bm{z}^i_{\ell}}\right)\geq 1+\text{exp}(-y^i(\bm{\beta}_\mathrm{x}^\top\bm{x}^i +\bm{\beta}_\mathrm{z}^\top\bm{z}) + \beta_0)\\
    \Leftrightarrow &\log \left(-1+\text{exp}(r_i + \lambda \sum_{\ell \in [m]} \delta_{\ell} \indicate{\bm{z}_{\ell} \neq \bm{z}^i_{\ell}} \right)\geq -y^i(\bm{\beta}_{\mathrm{x}}^\top\bm{x}^i +\bm{\beta}_{\mathrm{z}}^\top\bm{z} + \beta_0) \\
     \Leftrightarrow & y^i\left(\bm{\beta}_{\mathrm{x}}^\top\bm{x}^i + \beta_0\right) \geq -y^i \bm{\beta}_{\mathrm{z}}^\top\bm{z} - \log \left(-1+\text{exp}(r_i + \lambda \sum_{\ell \in [m]} \delta_{\ell} \indicate{\bm{z}_{\ell} \neq \bm{z}^i_{\ell}} \right).
\end{align*}
Therefore, the constraints indexed by $i \in [N]$ in problem \eqref{eq:convex reformulation} are equivalent to 
\begin{equation*} y^i\left(\bm{\beta}_{\mathrm{x}}^\top\bm{x}^i + \beta_0\right)\geq \max_{\bm{z} \in \mathcal{C}} -y^i \bm{\beta}_{\mathrm{z}}^\top\bm{z} - \log \left(-1+\text{exp}(r_i + \lambda \sum_{\ell \in [m]} \delta_{\ell} \indicate{\bm{z}_{\ell} \neq \bm{z}^i_{\ell}} \right)\end{equation*}
as described in Section \ref{sec:graph}, where the right hand side corresponds to problem \eqref{eq:constraint reformulation}.
Therefore, problem \eqref{eq:convex reformulation}
can be formulated as 
\begin{equation} \label{eq:convex reformulation revised}
\begin{array}{rl}
\underset{\lambda, \boldsymbol{r}, \bm{\beta}}{\text{min}} & \lambda \epsilon + \frac{1}{N} \sum_{i \in [N]} r_i \\[1.2em]
\text{s.t.} & y^i\left(\bm{\beta}_\mathrm{x}^\top\bm{x}^i + \beta_0\right)\geq \max_{\bm{z} \in \mathcal{C}} -y^i \bm{\beta}_{\mathrm{z}}^\top\bm{z} - \log \left(-1+\text{exp}(r_i + \lambda \sum_{\ell \in [m]} \delta_{\ell} \indicate{\bm{z}_{\ell} \neq \bm{z}^i_{\ell}} \right), \; \forall i \in [N] \\[3mm]
&  |\gamma_{j}^{-1} \beta_{\mathrm{x}j}| \leq \lambda, \; \forall j \in [n] \\[1em]
& \lambda \geq 0, \; \boldsymbol{r} \in \mathbb{R}^N, \; {\bm \beta} \in \mathbb{R}^{1+n+c}
\end{array}
\end{equation}

Next, We transform problem \eqref{eq:constraint reformulation} into a more tractable formulation. According to Lemma \ref{dp to path}, problem \eqref{eq:constraint reformulation} can be treated as a longest path problem, which can subsequently be reformulated as the following linear program.

\[
\begin{array}{ll}
    \max\limits_{\bm{a}^i \in \mathbb{R}^{|\mathcal{A}^i|}} & \sum\limits_{e \in \mathcal{A}^i} w^i(e) a_{e}^i \\[1mm]
    \quad \text{s.t.} & \sum\limits_{\substack{\{e \in \mathcal{A}^i | \\ t^i(e) = v\}}} a_{e}^i 
    - \sum\limits_{\substack{\{e \in \mathcal{A}^i | \\ s^i(e) = v\}}} a_{e}^i = 
    \begin{cases} 
    -1 & \text{if } v = (0, 0)\\ 
    1 & \text{if } v = (m+1, 0)\\
    0 & \text{otherwise}
    \end{cases}, \quad \forall v \in \mathcal{V}^i \\[9mm]
     &\bm{a}^i \geq \bm{0}
\end{array}
\]
where $a_e^i$ is a decision variable indicating if we select arc $e \in \mathcal{A}^i$. Since strong duality holds for linear programing problems, we get its dual problem:
\begin{equation*}
    \begin{array}{ll}
    \min\limits_{\bm{\mu}^i \in \mathbb{R}^{|\mathcal{V}^i|}} & -\mu_{(0,0)}^i + \mu_{(m+1,0)}^i \\[1mm]
    \quad \text{s.t.} & \mu_{t^i(e)}^i - \mu_{s^i(e)}^i \geq w^i(e), \; \forall e \in \mathcal{A}^i  
\end{array}
\end{equation*}
where $\bm{\mu}^i \in \mathbb{R}^{|\mathcal{V}^i|} $ are dual variables. 
We can convert problem \eqref{eq:convex reformulation revised} into 
\begin{equation*}
\begin{aligned}
& \underset{\lambda,\boldsymbol{r},\bm{\beta}, \bm{\mu}}{\text{min}}
& & \lambda \epsilon + \frac{1}{N}\sum_{i \in [N]} r_i \\
& \quad \text{s.t.}
& &    y^i(\bm{\beta}_{\mathrm{x}}^\top \bm{x}^i +\beta_0)\geq \min_{\bm{\mu}^i \in \mathbb{R}^{|\mathcal{V}^i|}} -\mu_{(0,0)}^i + \mu_{(m+1,0)}^i, \; \forall i \in [N]  \\
& & &  \mu_{t^i(e)}^i - \mu_
{s^i(e)}^i \geq w^i(e), \; \forall i \in [N], \, e \in \mathcal{A}^i\\
& & & |\gamma_{j}^{-1} \beta_{\mathrm{x}j}| \leq \lambda, \; \forall j \in [n] \\
& & & \lambda \geq 0, \; \boldsymbol{r} \in \mathbb{R}^N, \; {\bm \beta} \in \mathbb{R}^{1+n+c}
\end{aligned}
\end{equation*}

Instead of separately minimizing over $\bm{\mu}^i$,
 we can remove the minimization operator directly without changing the feasible region because the minimization problem ensures the existence of a feasible solution that satisfies certain constraints. Therefore, we can get
\begin{equation*}
\begin{aligned}
& \underset{\lambda,\boldsymbol{r},\bm{\beta}, \bm{\mu}}{\text{min}}
& & \lambda \epsilon + \frac{1}{N}\sum_{i \in [N]} r_i \\
& \quad \text{s.t.}
& &    y^i(\bm{\beta}_{\mathrm{x}}^\top \bm{x}^i +\beta_0)\geq -\mu_{(0,0)}^i + \mu_{(m+1,0)}^i, \; \forall i \in [N]  \\
& & &  \mu_{t^i(e)}^i - \mu_
{s^i(e)}^i \geq w^i(e), \; \forall i \in [N], \, e \in \mathcal{A}^i\\
& & & |\gamma_{j}^{-1} \beta_{\mathrm{x}j}| \leq \lambda, \; \forall j \in [n] \\
& & & \lambda \geq 0, \; \boldsymbol{r} \in \mathbb{R}^N, \; {\bm \beta} \in \mathbb{R}^{1+n+c}, \; \bm{\mu} \in \mathbb{R}^{\sum_{i \in [N]} |\mathcal{V}^i|}
\end{aligned}
\end{equation*}
\end{proof}

\vfill

\end{document}